\pdfoutput=1

\documentclass[11pt]{article}

\usepackage{ACL2023}

\usepackage{times}
\usepackage{latexsym}

\usepackage[T1]{fontenc}

\usepackage[utf8]{inputenc}

\usepackage{microtype}

\usepackage{inconsolata}

\usepackage{todonotes}

\usepackage[utf8]{inputenc} %
\usepackage[T1]{fontenc}    %
\usepackage{hyperref}       %
\usepackage{url}            %
\usepackage{booktabs}       %
\usepackage{multicol}
\usepackage{multirow}
\usepackage{amsfonts}       %
\usepackage{mathtools}
\usepackage{bm}             %
\usepackage{nicefrac}       %
\usepackage{microtype}      %
\usepackage{xcolor}         %
\usepackage{graphicx}
\usepackage{caption}
\usepackage{amsmath}
\usepackage{mwe}
\usepackage{float}
\usepackage{pdfpages}
\usepackage{subcaption}
\usepackage{booktabs}
\usepackage{graphicx}
\usepackage{cleveref}
\usepackage{mwe}
\usepackage{xspace}

\usepackage{tikz}
\usetikzlibrary{fit, patterns, matrix, calc}
\usepackage{pgfplots}
\pgfplotsset{compat=newest}

\newcommand{\model}{L_{prompt}}

\newcommand{\ie}{\textit{i.e.,}\@\xspace}

\newcommand{\etal}{\textit{et al.}\@\xspace}

\DeclareMathOperator*{\argmax}{arg\,max}

\newcommand{\nameA}{\textsf{Avg-Ens}\@\xspace}

\newcommand{\nameB}{\textsf{Vote-Ens}\@\xspace}

\newcommand{\numChosen}{50\xspace}
\newcommand{\numTuned}{1000\xspace}
\newcommand{\numEns}{50\xspace}

\newif\ifdraft
\drafttrue

\ifdraft
\newcommand{\franzi}[1]{{\color{purple}{\bf~F:~}#1}}
\newcommand{\adam}[1]{{\color{blue}Adam: #1}}

\else
\newcommand{\franzi}[1]{}
\newcommand{\adam}[1]{}

\fi

\title{On the Privacy Risk of In-context Learning}

\author{Haonan Duan, Adam Dziedzic, Mohammad Yaghini \\ \textbf{Nicolas Papernot}, \textbf{Franziska Boenisch}
\\
  University of Toronto \\
  Vector Institute \\
  }

\begin{document}
\maketitle
\begin{abstract}
Large language models (LLMs) are excellent few-shot learners. They can perform a wide variety of tasks purely based on natural language prompts provided to them. These prompts contain data of a specific downstream task---often the private dataset of a party, e.g., a company that wants to leverage the LLM for their purposes. We show that deploying prompted models presents a significant privacy risk for the data used within the prompt by instantiating a highly effective membership inference attack. We also observe that the privacy risk of prompted models exceeds fine-tuned models at the same utility levels. After identifying the model's sensitivity to their prompts---in form of a significantly higher prediction confidence on the prompted data---as a cause for the increased risk, we propose ensembling as a mitigation strategy. By aggregating over multiple different versions of a prompted model, membership inference risk can be decreased.

\end{abstract}

\section{Introduction}

Large language models (LLMs) exhibit strong %
capabilities for few-shot learning.
When provided with a natural language prompt in the form of %
a small number of examples for the specific context, the models can perform a myriad of natural language downstream tasks without modifications of their parameters~\citep{brown2020language,radford2019language}.
Prompting is more parameter and data-efficient than fine-tuning.
First, given a large number of parameters in LLMs, prompting boosts efficiency in downstream tasks~\citep{raffel2020exploring} without any adaptation of model parameters.
In contrast, fine-tuning requires retraining a significant fraction of parameters.
Second, it has been shown that prompting can leverage training data more efficiently than standard fine-tuning with a prompt being worth $\sim$100 data points~\citep{scao2021many}.%

The effectiveness of prompts is possible since the prompt data exhibit a significant effect on the LLMs' behavior~\citep{jiang2020can, shin2020autoprompt}.
This naturally raises the question of privacy risks.
Understanding privacy risks of prompting is of high importance since, in contrast to the large \textit{public} corpora used to pre-train the LLMs, the data used for prompting usually stems from a smaller \textit{private} downstream dataset.
Prior work has extensively studied the topic of memorization and privacy in LLMs~\citep{carlini2019secret,carlini2021extracting,zhang2021counterfactual}.
Yet, the considerations were limited to the data used for pre-training the LLMs~\citep{carlini2019secret,zhang2021counterfactual} or to fine-tune the model parameters~\citep{li2022large,yu2022differentially,zhang2022text}.
In contrast, we analyze how much privacy of the data used for prompting leaks from the deployed prompted LLM.
With our results, we are the first to show that prompted LLMs exhibit a high risk to disclose the membership of their private prompt data.

In our study, we focus on text generation models~\cite{radford2018improving,radford2019language} %
prompted with a proper template for any given downstream classification task. %
In this setup, we study privacy leakage through the lense of membership inference attacks (MIA)~\cite{carlini2022membership, shokri2017membership}---currently the most widely applied approach for estimating practical privacy leakage.
With access only to the probability vector output by the prompted LLM for a given input, we instantiate the MIA to determine whether this input was part of the prompt.
Our results suggest that data points used within the prompt are highly vulnerable to MIAs. %
Furthermore, in a controlled environment, we empirically evaluate the MIA-risk of prompting to the risk of fine-tuning with private data.
We find that prompted models are more than five times more vulnerable than fine-tuned models.

The severe vulnerability of the private prompt data and the fact that finding the high-performing prompts for a given downstream task requires significant human efforts and computing resourses ~\citep{zhou2023large} demand the design of protection methods.
Based on the observation that the prompted LLMs exhibit a significant higher prediction confidence on their prompted data---leading to the great success of MIA---we propose an effective defense:
We show that by ensembling over different prompted versions of an LLM, we can align the prediction confidence on prompt data (members), and other data (non-members) while achieving the same high prediction accuracy.
Obtaining such an ensemble of prompted models  is efficient since multiple well-performing prompts is already the by-products of our prompt-tuning and does not require additional steps.
We evaluate two concrete instantiations of prompt ensembling, namely \nameA and \nameB and quantify their effect on the risk of MIAs.
We show that our ensembling effectively reduce the success of MIA to close to random guessing.
Thereby, the privacy of the prompted data can be protected.

In summary, we make the following contributions:
\begin{itemize}
    \item We instantiate the first MIA on prompted LLMs and show that we can effectively infer the membership of the prompted data points with high success.
    \item We empirically compare the MIA risk of prompted and fine-tuned models in a controlled experimental environment and observe that the privacy risk of prompting significantly outperforms the one of fine-tuning.
    \item We demonstrate how to mitigate the privacy leakage we observed with prompt ensembling to a MIA-success rate of close to random guessing. 
\end{itemize}

\section{Background and Related Work}
\label{sec:background}

\subsection{Language Model Prompting}
\label{sub:prompts}

The success of LLMs such as the different versions of GPT~\citep{brown2020language,radford2018improving,radford2019language} and their exceptional few-shot learning capacities gave rise to prompt-based learning.
Without having to adapt any parameters, prompt-based learning leverages the capacities %
of LLMs and achieves similar downstream performance %
as full model fine-tuning~\citep{lester2021power,liu2021p}.
Therefore, it suffices to provide the model with a task-specific context in the form of a few examples, also called \textit{demonstrations}.
The prompt-based approach does improve computational and storage complexity over fine-tuning since no parameters of the underlying LLMs need to be updated and instead of having to save a fully fine-tuned model, only the required prompt has to be recorded. 
Prompts can be designed either manually by a human expert, or by an automated process~\citep{gao2020making,guo2022efficient,liu2021makes,shin2020autoprompt}. Our demonstrations come from the actual discrete vocabulary and we consider privacy leakage of the underlying data points -- sentences from the downstream tasks used for prompting.

\subsection{Memorization and Privacy Leakage in LLMs}
\label{sub:privacy_leakage}
LLMs are shown to memorize their training data which enables adversaries to extract this data when interacting with the model~\citep{carlini2022quantifying,ippolito2022preventing,kharitonov2021bpe,mccoy2021much,tirumala2022memorization,zhang2021counterfactual}.
It has, for example, been shown that GPT2 reproduces large passages with up to 1000 words of its original training data at inference time.
Additionally, privacy risks through memorization in fine-tuning have been observed by Mireshghallah~\etal in %
\cite{mireshghallah2022memorization}.
The only prior work around privacy leakage in prompt-based learning has used prompting to extract knowledge from LLMs and their underlying large (and often public) training corpora~\citep{davison2019commonsense,jiang2020can,petroni2019language}.
In our setup, we do not target the privacy of the LLM's training data---neither the original large corpora nor the data used to adapt the model through fine-tuning.
Instead, we are the first to study the privacy of data used to prompt an LLM to perform particular downstream task.

\begin{figure*}[t]
\centering
	\resizebox{0.8\textwidth}{!}{
		\begin{tikzpicture}
\def\dx{2.5}
\def\dy{2}

\node (inputs) [matrix of nodes, row sep=1.5mm, column sep=4mm,
column 1/.style={anchor=west, nodes={draw, rounded corners}},
column 2/.style={anchor=center, nodes={draw, circle, inner sep=0}}
] at (0, -0.2*\dy){
|[label=left:$x_1$]|``Malisa has the flu'' & $+$\\
|[label=left:$x_2$]|``Betty is feeling good'' & $+$\\
|[draw=none]|$\vdots$ \phantom{``Betty is feeling good''} & $+$\\
|[label=left:$x_n$]|``John Doe has cancer'' & $+$\\
};

\node (llm) [draw, minimum height=1.3*\dy cm] at (1.5*\dx, -0.1*\dy) {LLM};

\node (outs) [matrix of nodes, column sep=1mm] at (3.7*\dx, 0){
    $\dots$    & ``healthy''  & $\dots$  & ``sick''  & $\dots$  &                &                  &\tiny{``sick'',``healthy''}\\   
    $\dots$    &  0.1         & $\dots$  & 0.73       & $\dots$  &$\tilde{Y_1}$   & $\rightarrow$    & $[0.1, 0.73]$\\
    $\dots$    &  0.3         & $\dots$  & 0.1       & $\dots$  &$\tilde{Y_1}$   & $\rightarrow$    & $[0.3, 0.1]$\\
    $\vdots$   & $\vdots$     & $\vdots$ & $\vdots$  & $\vdots$ &                & $\rightarrow$    & $\vdots$\\
    $\dots$    & 0.01         & $\dots$  & 0.9       & $\dots$  &$\tilde{Y_n}$   & $\rightarrow$    & $[0.01, 0.9]$\\
};

\begin{scope}[inner sep=0, rounded corners]
\node[draw, fit=(outs-1-2)(outs-5-2)]{};
\node[draw, fit=(outs-1-4)(outs-5-4)]{};

\node[draw, fit=(outs-2-1)(outs-2-2)(outs-2-5), label={[label distance=2mm, right]right:}]{};
\node[draw, fit=(outs-3-1)(outs-3-2)(outs-3-5), label={[label distance=2mm, right]right:}]{};
\node[draw, fit=(outs-5-1)(outs-5-2)(outs-5-5), label={[label distance=2mm, right]right:}]{};
\end{scope}

\path let \p1=(llm) in node (prompt) [matrix of nodes, draw, rounded corners, anchor=west, column 1/.style={anchor=west}, nodes={font={\footnotesize}}] at (\x1-15, 1.3*\dy){
\node[align=left, font={\bfseries \footnotesize}]{Template};\\
\node[align=left]{$p_1$: ``John Doe has cancer''; ``sick''};\\
};

\foreach \r in {1, 2, 3, 4}{
    \draw[-latex] (inputs-\r-1.east) -- (inputs-\r-2.west);
    \draw[-latex] (prompt.south west) -- (inputs-\r-2.north);
    \draw[-latex] (inputs-\r-2) -- (llm);
}

\foreach \r in {2,2,3,5}{
    \draw[-latex] (llm) -- (outs-\r-1);
}

\node[draw, rounded corners, densely dotted, fit=(prompt)(inputs-4-2)(outs-5-6), label=below:{Prompted Target Model}]{};

\end{tikzpicture}    
}
\caption{\textbf{Setup for Prompting and MIA.} We prompt the LLM with different prompts (same template) for a downstream task. The LLM returns per-token probabilities for the next token in the sequence. The adversary has query access to the prompted LLM and obtains prediction probabilities for each possible target class of the downstream task.
}
\label{fig:setup}
\end{figure*}
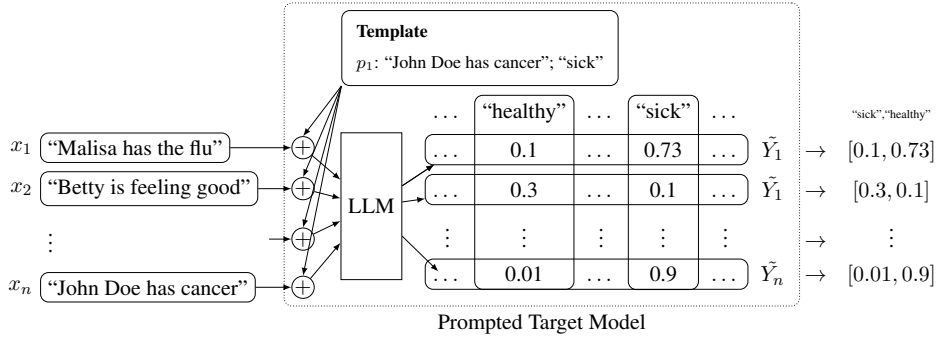

\subsection{Membership Inference Attacks}
\label{sub:mia}
When performing a membership inference attack (MIA)~\citep{carlini2022membership,shokri2017membership}, an adversary aims to determine whether a particular data point was used to train a given machine learning model.
The adversary usually has access only to the model's prediction outputs.
Membership inference attacks have been successful on a broad variety of machine learning models and domains, especially the vision~\citep{shokri2017membership,carlini2022membership} and language domain~\citep{shejwalkar2021membership,song2019auditing,carlini2021extracting}.
While a few prior works employ MIA to quantify memorization in LLMs~\cite{carlini2021extracting,oh2023membership,mireshghallah2022memorization}, they target the original large corpus training data or data used for fine-tuning the parameters of the models.
In contrast to them, we do not adapt the model parameters but freeze the entire LLM and design a prompt based on a small and private downstream dataset.
We evaluate MIA risk for the data points used within the prompt.

\subsection{Defending Against Membership Inference}
\label{sub:defenses}
Existing defense mechanisms against MIA can be divided into two main categories:
(i) empirical measures to reduce the adversary's attack success by either reducing model overfitting~\cite{chen2022relaxloss} or perturbing model outputs~\cite{muller2019does, jia2019memguard} and (ii) measures that rely on providing rigorous privacy guarantees according to Differential Privacy (DP)~\citep{dwork2006differential}.
These measure, for example, apply a DP stochastic gradient descent (DP-SGD)~\citep{abadi2016deep} while training a machine learning model.
However, in practice, DP significantly degrades performance of generative models~\citep{anil2021large} when not trained under very carefully chosen hyperparameters~\citep{li2022large}.
Therefore, none of the popular state-of-the-art LLMs is trained with DP. 
As a consequence, we focus our work on the  first category of defenses and propose ensembling multiple promoted models.
The only prior work using ensembles to defend against MIA is limited to small ML models for vision and tabular datasets, and requires a pre-processing over the entire training data at inference time to determine which of the ensemble models' training data did not contain the present data point~\cite{tang2022mitigating}.
We instead query all prompted models without additional pre-processing.

\section{Method}
\label{sec:method}

\subsection{Prompting for Downstream Classification}
\label{sub:prompting}
We focus on prompting pre-trained LLMs with the objective to perform a downstream classification task.
We denote the prompted model as $\model$.
Our prompts consist of tuples of demonstration sentences from the respective downstream task as prompt data, provided in a consistent template.
When applied to a specific input $x_i$, $\model$ predicts an $M$-dimensional probability vector $\tilde{y}_i$, with $M$ being the size of the vocabulary, where each component corresponds to the probability that the $\model$ assigns to the respective token for being the next token in the sequence $x_i$. 
Note that the output probabilities over all possible tokens are usually normalized such that $\sum_{m \in M} \tilde{y}_{i,m} = 1$.
Since we provide the model with demonstrations to solve a downstream classification task, the index with the highest values in $\tilde{y}_i$ should correspond to the token that represents the class label of the $x_i$.
For example on the input $x_i=$"The movie was great.", the highest probability should be for the token \textit{"positive"}, because this is the correct class label.
Given that the model is supposed to perform classification for a given downstream task, we assume that when querying $\model$ with $x_i$, not the entire $\tilde{y}_i$ has to be returned.
Instead, we are only interested in a subset of token-probabilities, namely for those tokens that correspond to classes in the respective downstream dataset.
We denote the reduced probability vector as $y_i$. 
Note that since $y_i$ only consists of a subset of the token probabilities from $\tilde{y}_i$, the probabilities in $y_i$ are unnormalized, \ie they do not necessarily add up to one, $\sum_{m \in M} y_{i,m} \leq 1$. We depict our setup in \Cref{fig:setup}.

\subsection{MIA Setup and Threat Model}
\label{sub:threat_model}
For our MIA, we assume an adversary with black-box access to the prompted model $\model$. 
This adversary can query $n$ text sequences $(x_1, \cdots, x_n)$ to $\model$ and obtains the output probability vectors $(y_1, \cdots, y_n)$.
Following a line of prior MIAs~\citep{jayaraman2021revisiting,yeom2018privacy}, we base our attack on the model's output probability at the token $y_{i,l}$ that corresponds to the correct target class label $l$.

\subsection{Prompt Ensembling}
\label{sub:prompt_ensembling}

\begin{figure}
	\resizebox{0.45\textwidth}{!}{
		\begin{tikzpicture}
\def\dx{2.5}
\def\dy{2}

\node (inputs) [matrix of nodes, row sep=1.5mm, column sep=4mm,
column 1/.style={anchor=west, nodes={draw, rounded corners}},
column 2/.style={anchor=center, nodes={draw, circle, inner sep=0}}
] at (0, 0){
|[draw=none]|$\vdots$ \phantom{``Betty is feeling good''} & $+$\\
|[label=left:$x_i$]|``Malisa has the flu'' & $+$\\
|[draw=none]|$\vdots$ \phantom{``Betty is feeling good''} & $+$\\
};

\path let \p1=(inputs-2-2) in node (llm) [draw, minimum height=1.3*\dy cm] at (1.5*\dx, \y1) {LLM};

\node (outs) [matrix of nodes, column sep=1mm] at (2.2*\dx, 0){
    $Y_{i,1}$ \\   
    $Y_{i,2}$ \\
    $\vdots$ \\
    $Y_{i,k}$  \\
};

\path let \p1=(outs-2-1) in 
    node (agg) [draw, rotate=90] at (\x1+30, \y1-10) {Aggregator} ;

\newcommand{\promptBlock}[4]{
    \path let \p1=(inputs-2-2) in node (pr#1) [matrix of nodes, draw, rounded corners, anchor=west, column 1/.style={anchor=west}, nodes={font={\footnotesize}}, , fill=white, fill opacity=0.75, text opacity=1] at (\x1+#2, 2.5cm+#3){
    \node[align=left, font={\bfseries \footnotesize}]{Template};\\
    \node[align=left]{$p_1$: #4};\\
    };
}

\promptBlock{1}{0}{0}{``Betty is feeling good''; ``healthy''}
\promptBlock{2}{2mm}{-2mm}{''David has brian cancer''; ``sick''}
\promptBlock{3}{4mm}{-4mm}{``Mary is feeling dizzy''; ``sick''}

\node[draw, rounded corners, densely dotted, fit=(pr1)(pr3)(inputs-2-2)(llm)(agg), label=below:{Prompted Target Model}]{};

\path let \p1=(agg) in node (targetOut) [] at (\x1+2.2cm, \y1) {$\bar{Y}_i$: ``sick''};

\foreach \r in {1, 2, 3}{
    \draw[-latex] (inputs-2-1.east) -- (inputs-\r-2.west);
    \draw[-latex] (pr\r.south west) -- (inputs-\r-2.north);
    \draw[-latex] (inputs-\r-2) -- (llm);
}

\foreach \r in {1,2,4}{
    \draw[-latex] (llm) -- (outs-\r-1);
    \draw[-latex] (outs-\r-1) -- (agg);
}

\draw[-latex] (agg) -- (targetOut);
\end{tikzpicture}    
	}
	\caption{\textbf{Ensemble of Prompted Models}. We ensemble multiple prompted models with disjoint data and the same template. The final prediction is an aggregate of outputs from each prompted model.
\label{fig:ensemble-schema}
}
\end{figure}
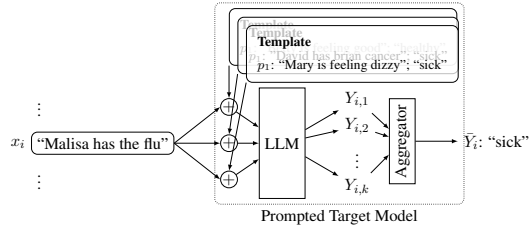

To mitigate the privacy risk, as exposed by prompt membership, we propose to aggregate the prediction probability vectors over multiple independent prompted models into an ensemble prediction, as shown in \Cref{fig:ensemble-schema}. We first tune $K$ prompted models $\model^{(1)}, \model^{(2)}, \dots, \model^{(K)}$. These $K$ models are prompted with disjoint training data. We then introduce two standard techniques to ensemble these prompted models ~\cite{jiang2020can,lester2021power} and refer to them as \nameA and \nameB.

In \nameA, we average the raw probability vectors of each of our $K$ prompted models $\model^{(1)}, \model^{(2)}, \dots, \model^{(K)}$.
Let $y_i^{(k)}$ be the output of our $k$th prompted model on input $x_i$.
The output of the ensemble $\model^{\text{\nameA}}$ on input $x_i$ is obtained as follows
\begin{align}
\model^{\text{\nameA}}(x_i) := \frac{1}{K} \sum_{k=1}^{K} \model^{(k)}(x_i)\text{.}
\end{align}

For \nameB, we rely on a majority vote of all the prompted models.
Therefore, we first obtain a single model's prediction on input $x_i$ as the token (class) from vocabulary $\mathcal{V}$ with the highest logit value as $\argmax(\model^{(k)}(x_i))$.
Let $n_v$ denote the number of prompted models that predict token $v$. 
Then, we return the token predicted by most models as
\begin{align}
\model^{\text{\nameB}}(x_i) :=\argmax_{v \in \mathcal{V}}\left(n_v\right)\text{.}
\end{align}

We do not evaluate the ensembling methods from Jiang~\etal~\cite{jiang2020can} that rely on (i) using the prompted model with the highest test accuracy as the output of the ensemble, or (ii) using a weighted average over the prompted models.
While (i) might yield utility improvements as shown in~\cite{jiang2020can}, it does not provide any privacy protection to the prompted model whose output is returned.
This is because the prediction of the ensemble still depends solely on a single model and thereby puts the privacy of that model at risk.
Since~\cite{jiang2020can} shows for (ii) that the weight concentrates on one single prompted model, the same impact on this model's privacy holds.

\paragraph{MIA on Ensembled Models.}
We also perform MIA on the ensembled models to study how ensembling mitigates the privacy risks.
For \nameA, we rely on the averaged output vector of the ensemble $y_i^{\text{\nameA}} = \model^{\text{\nameA}}(x_i)$, and extract the respective confidence value at the correct target class $y_{i,l}^{\text{\nameA}}$.
For \nameB, we count the number of prompted models that predict the target class $l$ and divide by the total number of models in the ensemble as $\frac{n_l}{K}$. 
Our empirical evaluation on the privacy risk mitigation through ensembled prompts is presented in \Cref{sub:evaluating_ensemble}.
\section{Experimental Evaluation}
\label{sec:experiments}

We experimentally study the MIA success on prompted LLMs and show that the prompted data exhibits a high vulnerability to MIAs.
Furthermore, we provide a comparison to the privacy risk of fine-tuning. We find that, at the same downstream accuracy, the privacy risk of prompt data in a prompted LLM surpasses the one of data used for fine-tuning. Finally, we demonstrate how ensembling the prediction of multiple prompted LLMs can effectively reduce the MIA risk close to random guessing.

\subsection{Experimental Setup}
\label{sub:exp_setup}

\begin{table}[t]
\centering
\small
\begin{tabular}{lcccccc}
\toprule
       & $N_\text{train}$ & $N_\text{test}$ & \# Classes & $min_\text{acc}$ & $max_\text{acc}$ \\ \midrule %
agnews & 12000             & 7600          & 4  &  0.65           &     0.83      \\ %
cb     & 250               & 56            & 3   &  0.60              &   0.73     \\ %
sst2   & 6920              & 1821          & 2     & 0.78              &   0.88     \\
trec    & 5452              & 500           & 6     & 0.40              & 0.59      \\
\bottomrule
\end{tabular}
\caption{\textbf{Evaluation Datasets.} Summary of the datasets and utility overview. We depict the number of training ($N_\text{train}$) and test data points ($N_\text{test}$), and the number of classes (\# Classes) in the task. Additionally, among the 50 selected best prompted LLMs, we report the span of their respective validation accuracies between the worst performing $min_\text{acc}$ and the best performing $max_\text{acc}$. The validation accuracy is used to find the best \numChosen prompted models among the \numTuned generated promoted models.}
\label{tab:dataset}
\end{table}

We prompt GPT2~\citep{brown2020language}\footnote{If not specified differently, we use GPT2-xl taken from HuggingFace (1.5 billion parameters).} to solve four standard downstream text classification tasks, namely \textit{agnews}~\citep{zhang2015character}, \textit{cb}~\citep{de2019commitmentbank}, \textit{sst2}~\citep{socher2013recursive} and \textit{rte}~\citep{dagan2006pascal}.
We document details of the datasets in~\Cref{tab:dataset}. 
Note that 20\% of the training data sets serve us as separate validation sets.

\paragraph{Tuning.} 
Our procedures for prompt tuning follow Zhao~\etal~\cite{zhao2021calibrate} unless otherwise specified.
Unlike them, we also return the probabilities for class labels whose corresponding tokens do not fall under the top 100 tokens.
This enables us to perform our MIA in a unified way over all model outputs.

Since the performance of prompted models is known to suffer from instability~\citep{dodge2020fine,zhangrevisiting}, we prompt the model \numTuned times with different 4-shot data from the downstream dataset.
We then keep the \numChosen best-performing prompts with disjoint data based on validation accuracy.\footnote{This type of prompt engineering corresponds to choosing the model with the best hyperparameters in standard training or fine-tuning.
}
The range of validation accuracies of the best selected \numChosen prompted models is reported in~\Cref{tab:dataset}.

\paragraph{MIA.} To evaluate our MIAs, we consider the data points used within the prompt of a model as members and all remaining training data points from the respective dataset as non-members.
This skewed distribution between members and non-members corresponds to a realistic scenario where only a small proportion of the candidate data targeted by the adversary are members~\cite{jayaraman2021revisiting}.
If not stated otherwise, we perform MIA on the unnormalized probability outputs of the prompted LLMs at the data point's correct target class.
To quantify the success of our attack, we report the AUC score as well as the true-positive rate (TPR) at low false-positive rates (FPRs). A successful MIA should have a high AUC score as well as a high TPR at low FPRs. 

\begin{figure}[t]
    \centering
    \includegraphics[width=0.475\textwidth]{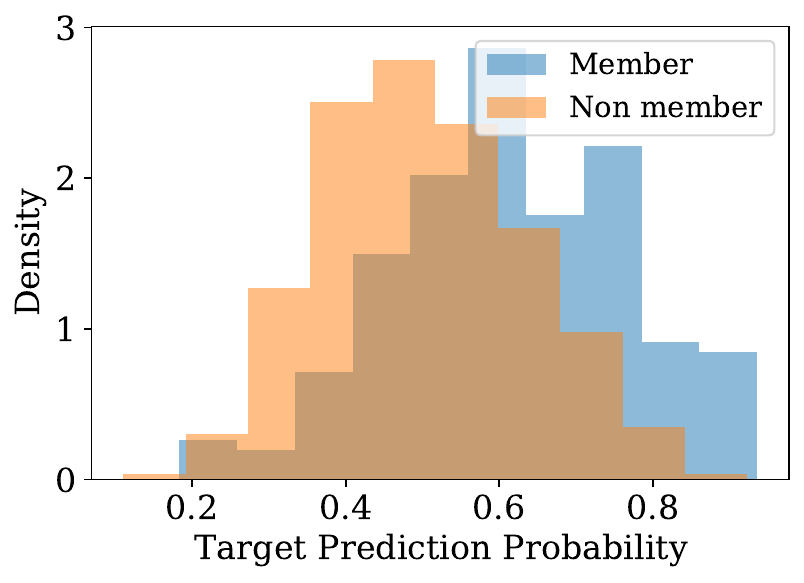}
    \caption{\textbf{Prediction Probability at Target Class (sst2).} 
    We plot output prediction probability for the target class for member and non-member data points of the prompt in the prompted LLM. 
    We find that the LLMs outputs for the prompt's member data is significantly higher than for non-member data points.
}%
    \label{fig:argnews_probs}
\end{figure}

\subsection{Success of Membership Inference Attack}
\label{sub:evaluating_MIA}

\begin{figure*}
        \begin{subfigure}[b]{0.235\textwidth}
            \centering
            \includegraphics[width=\textwidth]{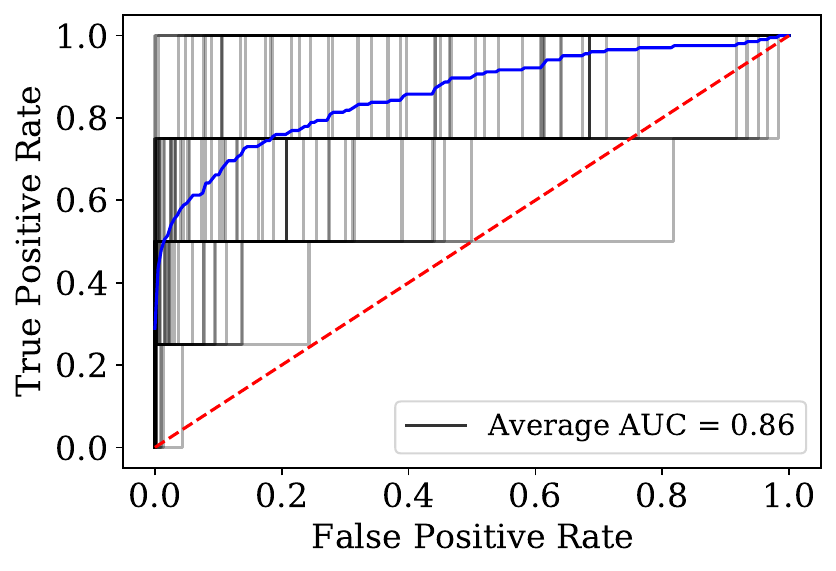}
            \caption[]%
            {{\small agnews}} 
            \label{fig:auc_argnews}
            \label{fig:mean and std of net14}
        \end{subfigure}
        \hfill
        \begin{subfigure}[b]{0.235\textwidth}  
            \centering 
            \includegraphics[width=\textwidth]{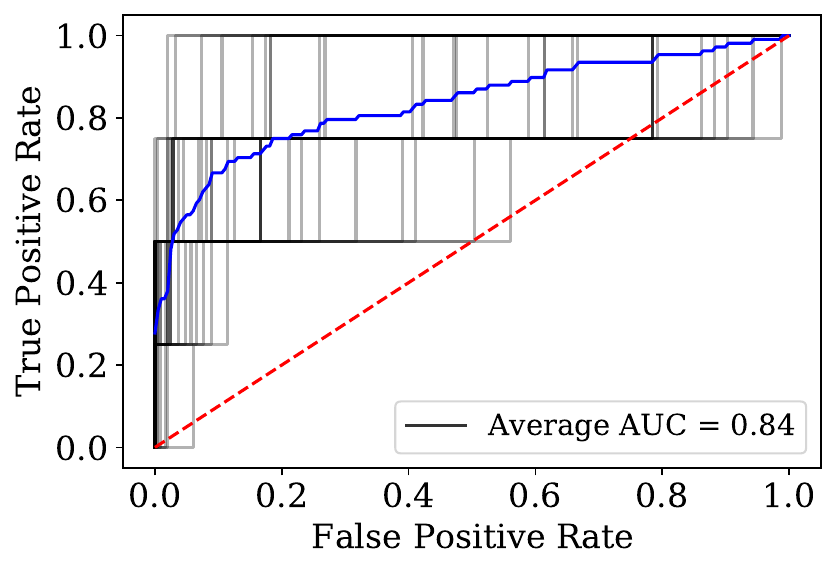}
            \caption[cb]%
            {{\small cb}}    
        \end{subfigure}
        \hfill
        \begin{subfigure}[b]{0.235\textwidth}   
            \centering 
            \includegraphics[width=\textwidth]{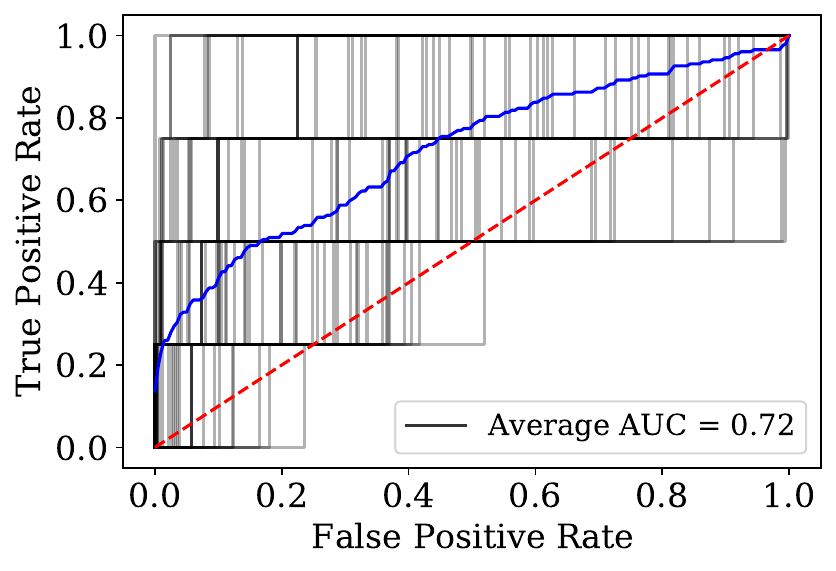}
            \caption[sst2]%
            {{\small sst2}}    
        \end{subfigure}
        \hfill
        \begin{subfigure}[b]{0.235\textwidth}   
            \centering 
            \includegraphics[width=\textwidth]{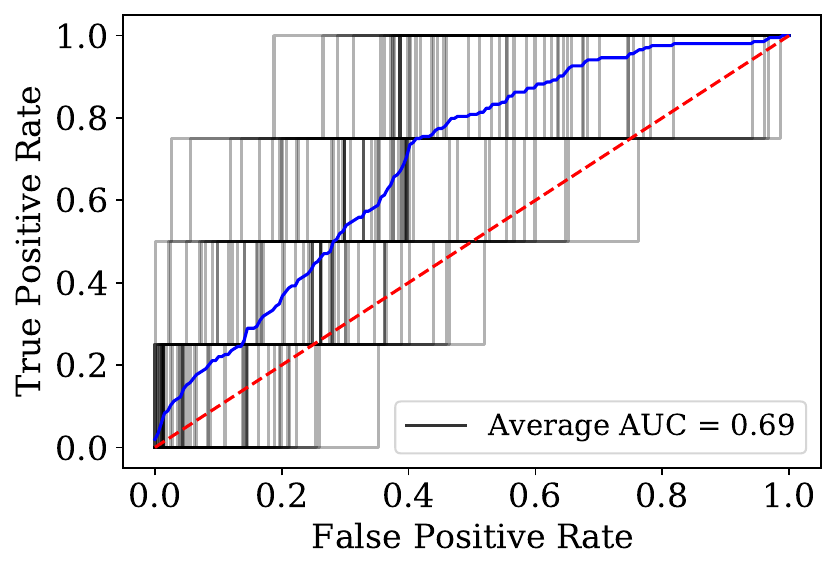}
            \caption[trec]%
            {{\small trec}}    
        \end{subfigure}
        \hfill 
        \caption{\textbf{MIA risk over all Datasets.} We depict the AUC-ROC curves over all datasets. The red dashed line
represents the MIA success of random guessing. Each gray line corresponds to a prompted model with its four member data points. Due to the small number of member data points (4), our resulting TPRs can only be 0\% 25\%, 50\%, or 100\% which leads to the step-shape of the gray curves.
        The reported average AUC-score is calculated as an average over the individual prompted models (gray lines)' AUC score.
        Additionally, for visualization purposes, we average the gray lines over all prompted models and depict the average as the blue line.
        We use \numChosen prompted models in this experiment.
        } 
        \label{fig:mi_attack}
    \end{figure*}

\begin{table*}[t]
\centering
\small
\begin{tabular}{lccccccc}
\toprule
         & \multicolumn{2}{c}{FPR=$1\mathrm{e}{-3}$} & \multicolumn{2}{c}{FPR=$1\mathrm{e}{-2}$} & \multicolumn{2}{c}{FPR=$1\mathrm{e}{-1}$} &   \\ 
         & Prompts & Fine-tuning & Prompts & Fine-tuning & Prompts & Fine-tuning \\
         \midrule
agnews & $0.222 \pm 0.212$ & $0.001 \pm 0.001$ & $0.433 \pm 0.281$ & $0.011 \pm 0.005$ & $0.661 \pm 0.253$& $0.105 \pm 0.010$  \\
cb   & $0.272 \pm 0.204$ & $0.051 \pm 0.071$ & $0.382 \pm 0.236$ & $0.111 \pm 0.120$ & $0.632 \pm 0.212$& $0.325 \pm 0.181$ \\ 
sst2  &  $0.137 \pm 0.187$& $0.002 \pm 0.003$ & $0.225 \pm 0.206$ & $0.018 \pm 0.009$ & $0.402 \pm 0.297$ & $0.167 \pm 0.0312$\\
trec & $0.019 \pm 0.067$ & $0.003 \pm 0.012$& $0.049 \pm 0.091$ & $0.023 \pm 0.038$& $0.221 \pm 0.201$ & $0.258 \pm 0.102$&  \\
\bottomrule
\end{tabular}
\caption{\textbf{TPR at at Different FPRs for Prompts and Fine-Tuning.} We report the TPR of our MIA at different low FPRs. 
The large standard deviation results from the small number of member data points (4).
We only consider FPRs down to $1\mathrm{e}{-3}$ which is larger than in \cite{carlini2022membership} which considers FPRs down to $1\mathrm{e}{-5}$.
This is because we operate on much smaller datasets where we cannot obtain such small fractions.
}
\label{tab:tprs_at_fprs}
\end{table*}

\begin{figure}[t]
    \centering
    \includegraphics[width=0.475\textwidth]{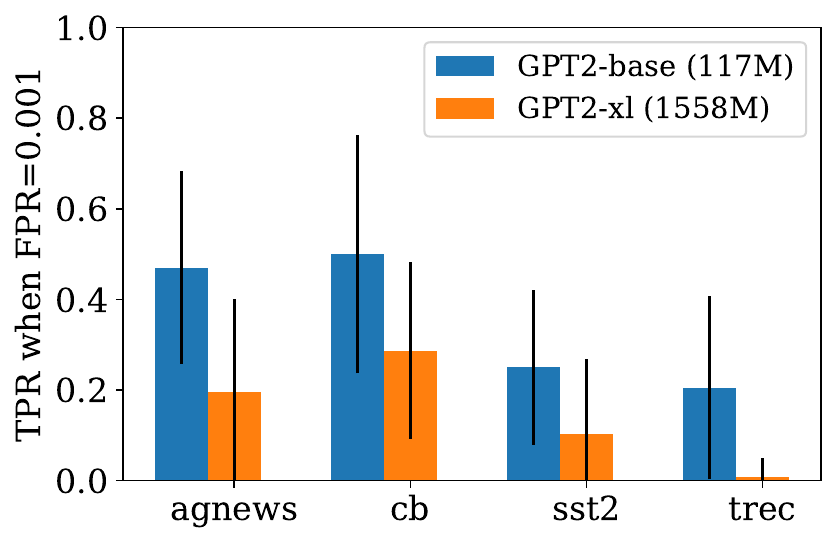}
    \caption{\textbf{Impact of Model Size on Membership Risk.} We report the TPR at FPR $1e-3$ for GPT2-base and GPT2-xl (117M vs 1.5B parameters). 
For fair comparison, we tune \numTuned prompts for both architectures, keep the best \numChosen for GPT2-base, and for GPT2-xl, we keep the \numChosen prompts that yield  validation accuracy closest to the one of GPT2-base.
We observe that larger models leak less private information about their prompts.
All results are obtained on the sst2 dataset.
}%
    \label{fig:model_size}
\end{figure}

We first analyze the probability output by the prompted LLM for the correct target class between member and non-member data points.
\Cref{fig:argnews_probs} shows for the sst2 dataset that the prediction outputs for non-members are overall lower than for members.
Similar results can be observed on all evaluation datasets, see \Cref{fig:distribution_probs} in \Cref{app:additional_results}.

This difference leads to a high MIA risk for the prompted data points as we show in \Cref{tab:tprs_at_fprs} and \Cref{fig:mi_attack}.
For example, on the sst2 dataset, on an FPR of $1e-3$, we observe a TPR of $0.137 \pm 0.187$, and an average AUC of $0.72$.
Note that the current most powerful MIA for supervised classification~\cite{carlini2022membership} obtains the same high AUC ($0.72$) score on the CIFAR10 dataset only by fully training 256 additional shadow models---a significant computational overhead we do not face.

\paragraph{Membership Risk is Higher on Smaller Models.}
We evaluate the impact of underlying LLMs' size on the vulnerability to MIAs against their prompted data. 
In this comparison, we focus on GPT2-base vs GPT2-xl. 
GPT2-base has 117M parameters, while GPT2-xl has 1.5B. 
For a fair comparison between the different models' vulnerability, we control the downstream performance of two models. 
Therefore, for GPT2-xl, we again generate \numTuned prompted models.
Among those, we keep the \numChosen prompts that lead to a performance close to the validation accuracy of the best prompted GPT2-base.
More precisely, we choose the \numChosen prompts for GPT2-xl that have a validation accuracy in the range of the \numChosen best models of GPT2-base. 
\Cref{fig:model_size} depicts the membership risk of prompted models of different sizes by depicting the TPRs at a FPR of 0.001. 
We find that GPT2-base consistently yields higher TPRs (\ie higher membership risk) than GPT2-xl across different datasets. 
We hypothesize that this disparate vulnerability is caused by larger models' better generalization capacity.
Larger models, when prompted with a few examples, due to their better generalization, have a smaller difference in output distribution between member and non-member data points.

\paragraph{Normalizing Prediction Probabilities.}
As we detail in \Cref{sub:prompting}, the prediction probabilities of the prompted model do not sum up to one since they are a only a small subset of all possible output tokens (whose total prediction probability sums up to one).
We evaluate how normalizing the model's output probabilities over all possible target classes in the downstream task influences the risk of MIA.
We depict our results in \Cref{fig:mi_attack_normalized} in \Cref{app:additional_results}.
The evaluation does not yield a consistent trend regarding the overall AUC among the datasets: while for argnews and trec the average AUC is similar with and without normalization, for cb and sst2, the raw outputs yield higher AUC. %
These results suggest that attackers can also perform successful MIA when the prediction outputs are processed in different ways---as it can happen when the prompted models are deployed behind some API.

\begin{figure}[t]
    \centering
    \includegraphics[width=0.475\textwidth]{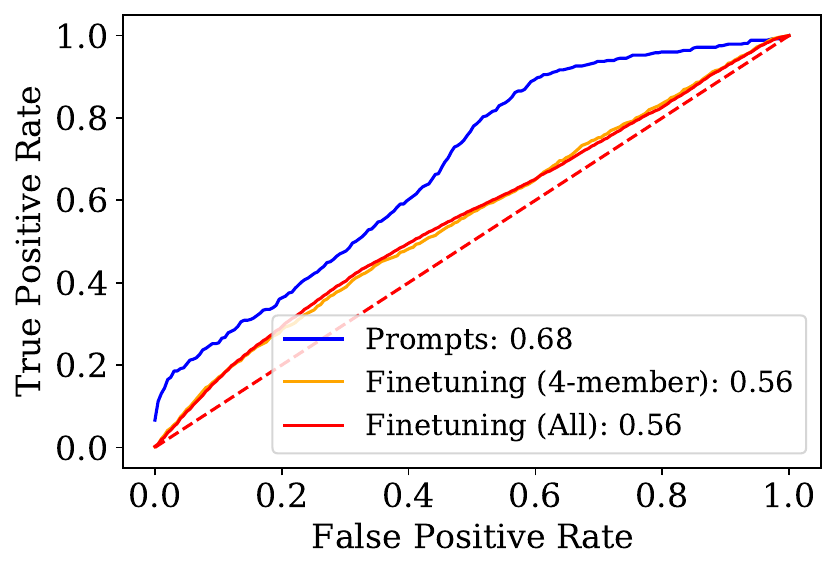}
    \caption{\textbf{Privacy Leakage in Fine-Tuning vs. Prompting (sst2).} 
    We plot the membership risk of our MIA on prompted and fine-tuned models given similar downstream performance.
    For fine-tuning, we evaluate MIA risk in two different ways to avoid the influence of different training set size. 
    The red dashed line represents the MIA success of random guessing.
    The results show that prompts are much more vulnerable to MIA than fine-tuning.
     Results of more datasets can be found in Figure \ref{fig:append_fine_tune_auc}.  
    }%
    \label{fig:fine-tune-auc}
\end{figure}

\begin{figure*}[t]
\begin{subfigure}[b]{0.235\textwidth}
            \centering
            \includegraphics[width=\textwidth]{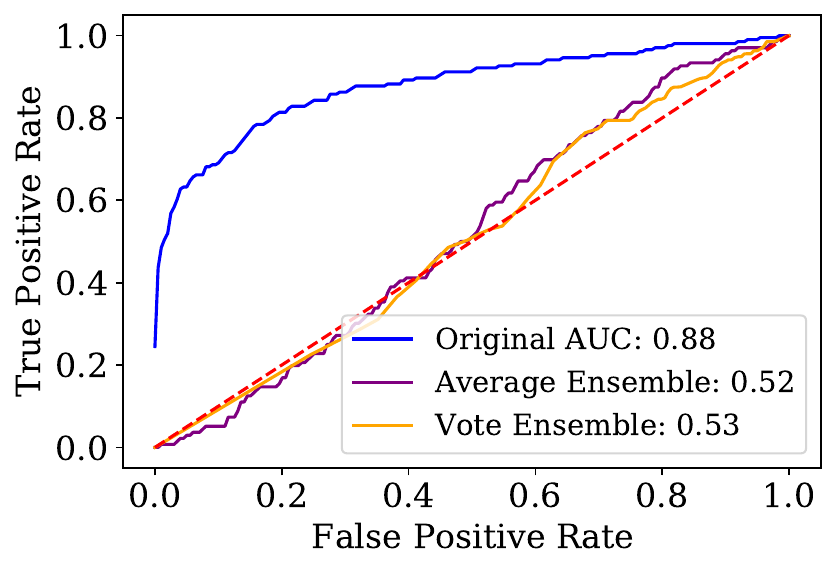}
            \caption[]%
            {{\small agnews}}    
            \label{fig:mean and std of net14}
        \end{subfigure}
        \hfill
        \begin{subfigure}[b]{0.235\textwidth}
            \centering
            \includegraphics[width=\textwidth]{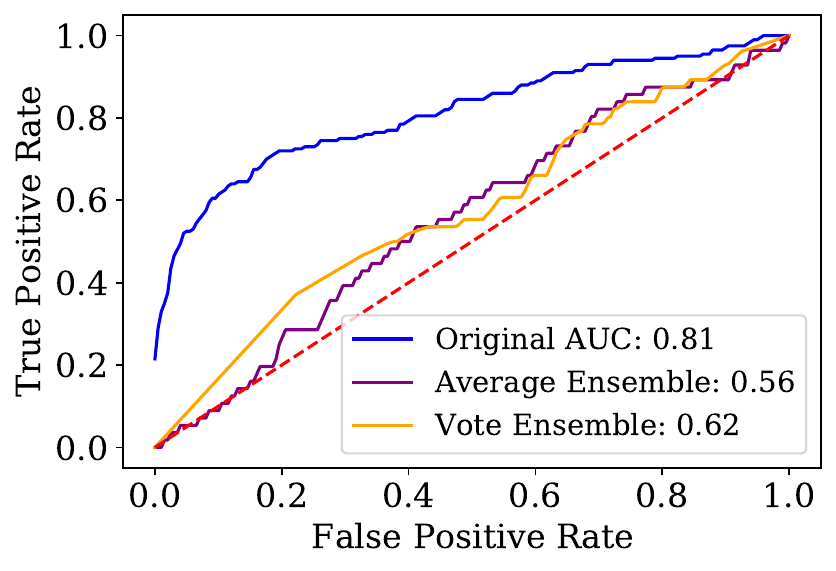}
            \caption[]%
            {{\small cb}}    
            \label{fig:mean and std of net14}
        \end{subfigure}
        \hfill
        \begin{subfigure}[b]{0.235\textwidth}  
            \centering 
            \includegraphics[width=\textwidth]{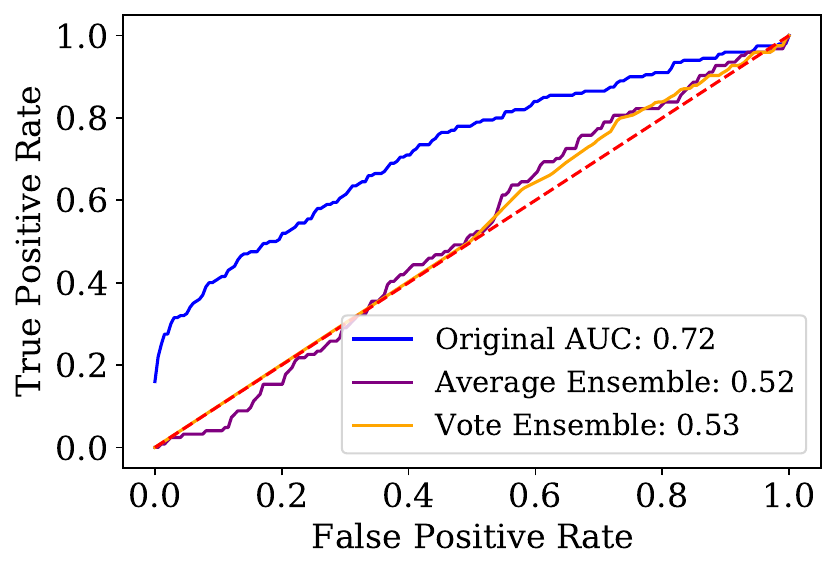}
            \caption[cb]%
            {{\small sst2}}    
            \label{fig:mean and std of net24}
        \end{subfigure}
        \begin{subfigure}[b]{0.235\textwidth}   
            \centering 
            \includegraphics[width=\textwidth]{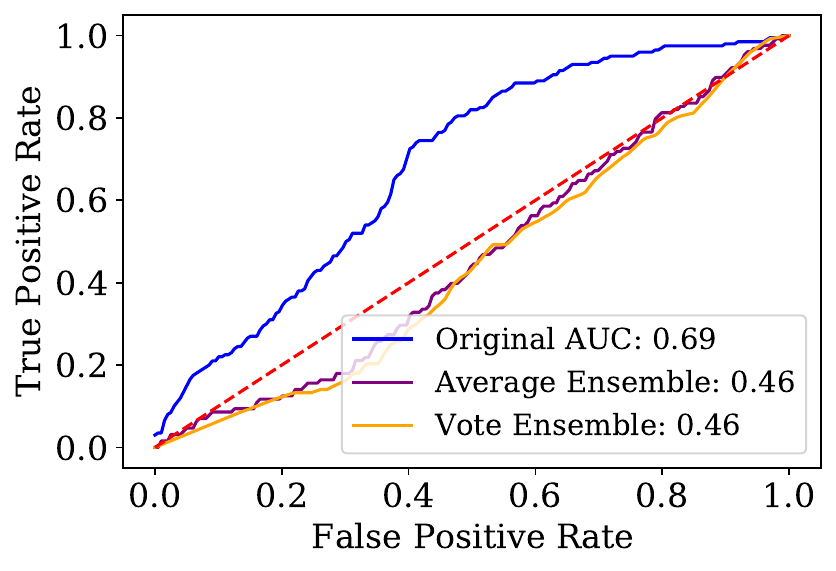}
            \caption[sst2]%
            {{\small trec}}    
            \label{fig:mean and std of net34}
        \end{subfigure}
        \hfill
        \caption{\textbf{Defense Method via Ensembling.} We depict the AUC-ROC curves over 4 datasets for our two ensembling defense methods (average, \nameA, and vote ensembles, \nameB) and compare it with the attack against the undefended model (blue solid line). The red dashed line represents random guessing. We find that ensembling effectively mitigates the threats of MIAs. }
        \label{fig:defense}
    \end{figure*}

\subsection{Prompting Leaks more Privacy than Fine-Tuning}
In this section, we compare the privacy leakage of prompting with fine-tuning.

\paragraph{Fine-Tuning Setup.}
Over all our experiments, we fine-tune only the last layer of GPT2 and a classification head. 
We fine-tune the model for $500$ epochs, and use the checkpoint with the highest validation accuracy during tuning.
For a controlled comparison between fine-tuning and prompting, our fine-tuned model's validation accuracy should roughly match the one of our prompted models.

Therefore, we first identify the number of data points needed for each downstream dataset to yield comparable validation accuracy.\footnote{Fine-tuning usually requires more data points than prompting~\citep{le2021many}.}
We run 100 fine tuning runs for each combination of the number of training data points ($4, 8, 32, 64, 128,256$) and learning rates ($1e-4, 1e-5, 1e-6$).
The number of data points needed and the corresponding learning rates are detailed in the table below:
\begin{table}[h]
    \footnotesize
    \centering
    \begin{tabular}{lccc}
    \toprule
      Dataset   &  \#Data Points & Learning Rate & Acc. \\
      \midrule
      agnews  & $512$ & $1e-5$ & 0.74\\
      cb  &$16$ & $1e-4$ & 0.68\\
      sst2 &$5536$&$1e-5$& 0.74\\
      trec &$32$& $1e-4$ &0.52\\
      \bottomrule
    \end{tabular}
    \caption{\textbf{Learning Parameters for Fine-Tuning.} We present the number of data points used for fine-tuning, the learning rates, and the resulting validation accuracies of our fine-tuning for all dataset.}
    \label{tab:fine-tuning-setup}
\end{table}

Note that for sst2, we were not able to meet the prompted models' validation accuracy (\Cref{tab:dataset}) even using the whole training dataset. 
Therefore, we compare with weaker prompts that yield accuracy between $0.72$ and $0.76$---instead of our \numChosen best selected ones.

\paragraph{MIA Evaluation Setup.}
Due to the different training set size in prompting and fine-tuning, for a fair comparison, we evaluate MIA for fine-tuned models in two ways:
(1) Following the setup for prompted models we select a different 4-tuple of members and evaluate against all the non-members from the validation set.
This procedure is repeated five times and we report the average over all resulting curves ROC curves and the average AUC.
(2) Following the standard setup for MIA~\cite{shokri2017membership} , we evaluate all the members together against the non-members and present the resulting ROC curve and AUC score.

\paragraph{Results.}
We present the MIA of fine-tuned models in Table \ref{tab:tprs_at_fprs} and Table \ref{fig:fine-tune-auc}. 
Our findings highlight that prompting yields higher privacy risks than fine-tuning under similar downstream performance. 
For example, at an FPR of $1e-3$, the average TPR for prompting is at least five times higher than for fine-tuning across all datasets.

\begin{table}[t]
\centering
\small
\begin{tabular}{lccc}
\toprule
       & $mean_\text{acc}$ & \nameA & \nameB  \\ \midrule %
agnews &  0.734            & 0.822 & 0.794            \\ %
cb     &  0.625            & 0.696 & 0.696             \\ %
sst2   &   0.854            & 0.904 & 0.908          \\
trec    &  0.406             & 0.520 & 0.500           \\
\bottomrule
\end{tabular}
\caption{\textbf{Test Accuracy of Ensembles.} We depict the validation accuracies of our initial prompted models (mean over all \numEns models) and the validation accuracies of our ensembling methods \nameA and \nameB. 
}
\label{tab:performance_ensembling}
\end{table}

\subsection{Ensembling Mitigates Privacy Risks}
\label{sub:evaluating_ensemble}

Finally, we experimentally evaluate the impact of our two ensembling approaches on the membership risk.
We report performance of our ensembles on the test data in \Cref{tab:performance_ensembling} and observe that both approaches perform equally well.

To study the impact on privacy risk, we first analyze the distribution of member and non-member data points' probability at the target class for \nameA.
\Cref{fig:distribution_ensembling} in \Cref{app:additional_results} highlights that through ensembling, the distributions for member and non-member probabilities become much more similar.
This also reflects in reduced membership risk as we depict in \Cref{fig:defense}. We find that for both methods, the attack curve after ensembling is close to random guessing (red line) across all datasets. 
Similar results are obtained with \nameB as we show in \Cref{fig:distribution_vote} and \Cref{fig:mi_attack_vote} in \Cref{app:additional_results}.

\begin{figure}[t]
    \centering
    \includegraphics[width=0.475\textwidth]{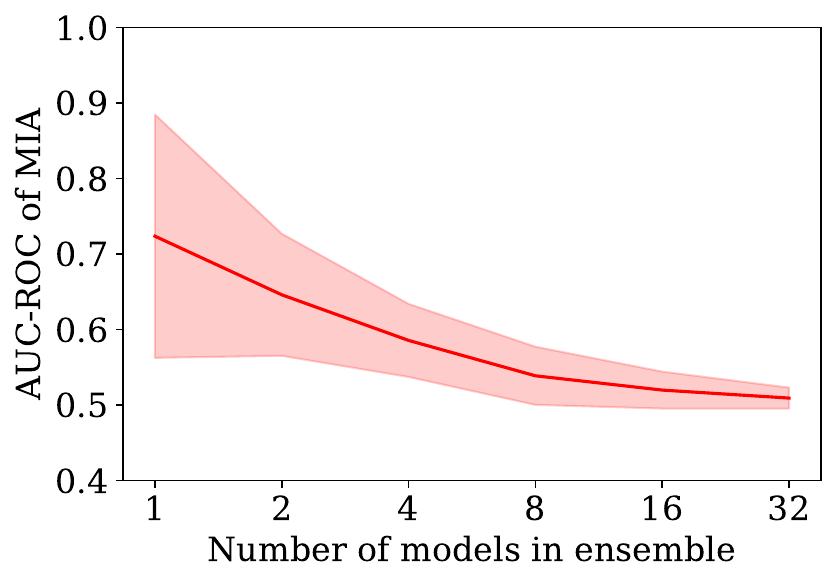}
    \caption{\textbf{MIA Risk vs Number of Models in an Ensemble (sst2).} We plot the membership risk in form of the AUC score 
    of MIA while we vary the number of teachers in ensembling. Results of other datasets can be found in Figure \ref{fig:append_number_of_teachers}. We observe that with more data used for ensembling, the lower risk of MIA (in terms of AUC and its variance).
    }%
    \label{fig:number_of_teachers}
\end{figure}

Finally, we evaluate the influence of the number of prompted model in the ensemble on the resulting membership risk. 
\Cref{fig:number_of_teachers} highlights that with an increasing number of prompted models in the ensemble, privacy risk decreases. 
This effect results from the fact that averaging over more models generally implies smaller influence of one particular model. 
However, there is a trade-off between increased inference times and the decreased privacy costs of using larger ensembles.
Our \Cref{fig:append_number_of_teachers} in \Cref{app:additional_results} suggest that using as little as $16$ teachers could reduce the average MIA for all datasets below $0.55$, \ie close to random guessing.

\section{Conclusions}
\label{sec:conclusion}
We are the first to show that prompted LLMs exhibit a high risk to disclose the membership of their private prompt data. %
To determine the membership of a data point, it is sufficient for an attacker to analyze the model's prediction confidence at the target class.
When comparing the privacy risk of prompted models with standard fine-tuning, 
we observe that prompts exhibit a higher privacy leakage than fine-tuning.
However, there are many advantages of prompts over fine-tuning. For example, instead of storing multiple versions of the whole fine-tuned model per downstream task, the underlying LLMs stay intact while only the prompt changes to implement different tasks.
Thus, to mitigate privacy risks for prompts, we propose ensembling over multiple prompted models.
We experimentally validate that this approach reduces the membership risk of the prompt data.
An interesting observation is that privacy leakage also decreases with the increasing number of language model parameters. %
This suggests a general trend that the prompt data become less vulnerable to privacy risks with a better generalization of the models.

\section*{Limitations}

While our ensembling approach empirically mitigates the risk of MIAs against prompted LLMs, we acknowledge that the approach does not provide rigorous privacy guarantees.
Future effort should be put into extending our approach to implement, for example, differential privacy~\cite{dwork2006differential}.

Furthermore, we acknowledge that our ensembling approach creates computational overhead since inference needs to be run with multiple prompts instead of a single one. This disadvantage can be reduced by running inference over all the prompts in a batch.

In this work, we solely consider discrete prompts due to their popular usage. 
There exist also \textit{soft prompts}~\citep{qin2021learning, zhong2021factual} that are optimized sequences of continuous task-specific input vectors. They are not tied to embeddings from the vocabulary. 
The privacy leakage of soft prompts and designing potential defenses will be addressed in our future work.

Finally, due to the cost associated with access to GPT-3, we limit our empirical evaluations to GPT-2 which is available as an open-source model.
To reduce potential biases that might arise through this limitation, we evaluated on different versions of GPT-2, including GPT2-xl, which has >1.5B parameters.

\bibliographystyle{plainnat}
\bibliography{main}

\appendix
\section{Broader Impact and Ethics Statement}
Prompting is on the way of becoming a highly prominent paradigm of using LLMs---which makes assuring the privacy of the prompt data an urgent need.
We present an empirical yet efficient mitigation of privacy risks but we acknowledge that this approach does not yield formal privacy guarantees.
Therefore, we encourage model owners to use our MIA as a tool to to empirically evaluate the privacy of their prompted model, or their ensemble of prompted models, before deployment.
A high MIA score should galvanize the model owners to implement additional protection before the deployment. 

By relying purely on open-source LLMs and public open source datasets in our experimental evaluation, we make sure that the result reported in the current work do not harm individuals' privacy.
We also recognize the importance of transparency in machine learning research, and we have made efforts to provide clear explanations of our methods and results, and provide additional experimental results on multiple datasets in the Appendix.
\newpage
\section{Additional Experimental Results} %
\label{app:additional_results}

\begin{figure*}
        \begin{subfigure}[b]{0.225\textwidth}
            \centering
            \includegraphics[width=\textwidth]{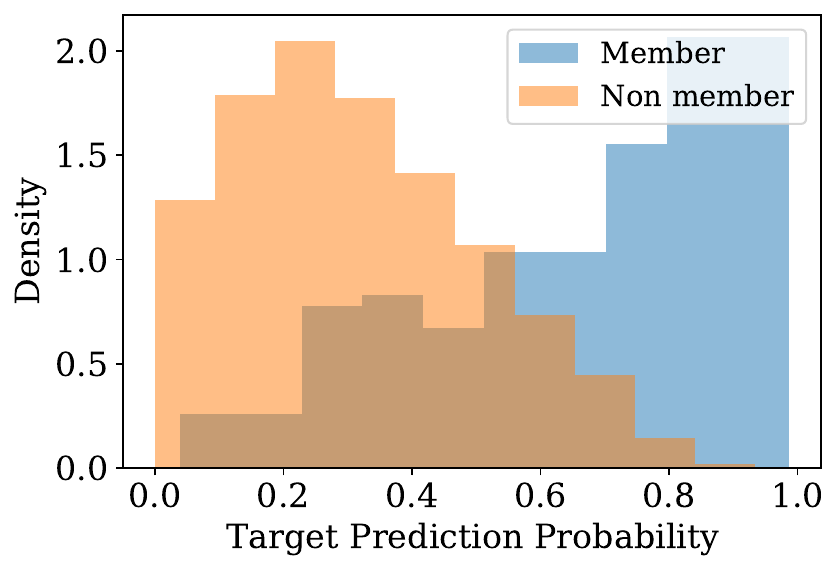}
            \caption[]%
            {{\small agnews}}    
            \label{fig:mean and std of net14}
        \end{subfigure}
        \hfill
        \begin{subfigure}[b]{0.225\textwidth}  
            \centering 
            \includegraphics[width=\textwidth]{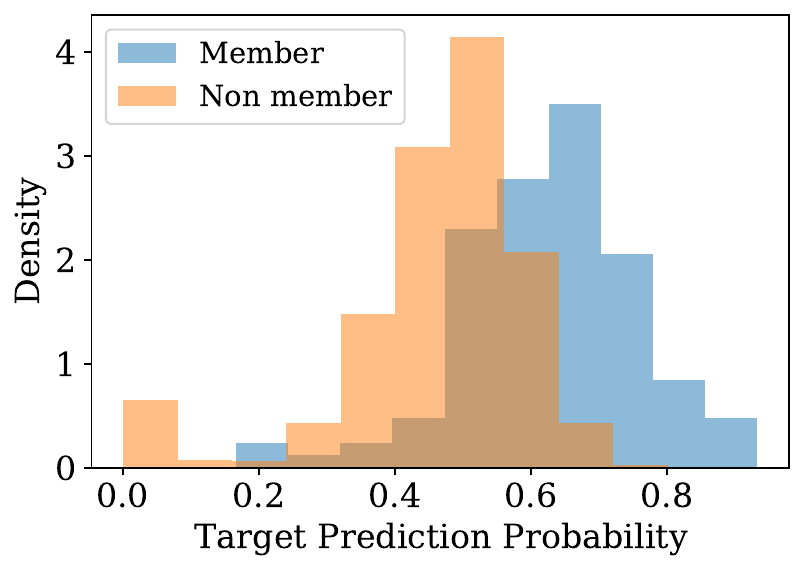}
            \caption[cb]%
            {{\small cb}}    
            \label{fig:mean and std of net24}
        \end{subfigure}
        \hfill
        \begin{subfigure}[b]{0.225\textwidth}   
            \centering 
            \includegraphics[width=\textwidth]{figure/sst2/histogram_raw_sst2_gpt2-xl_4_shot.pdf}
            \caption[sst2]%
            {{\small sst2}}    
            \label{fig:mean and std of net34}
        \end{subfigure}
        \hfill
        \begin{subfigure}[b]{0.225\textwidth}   
            \centering 
            \includegraphics[width=\textwidth]{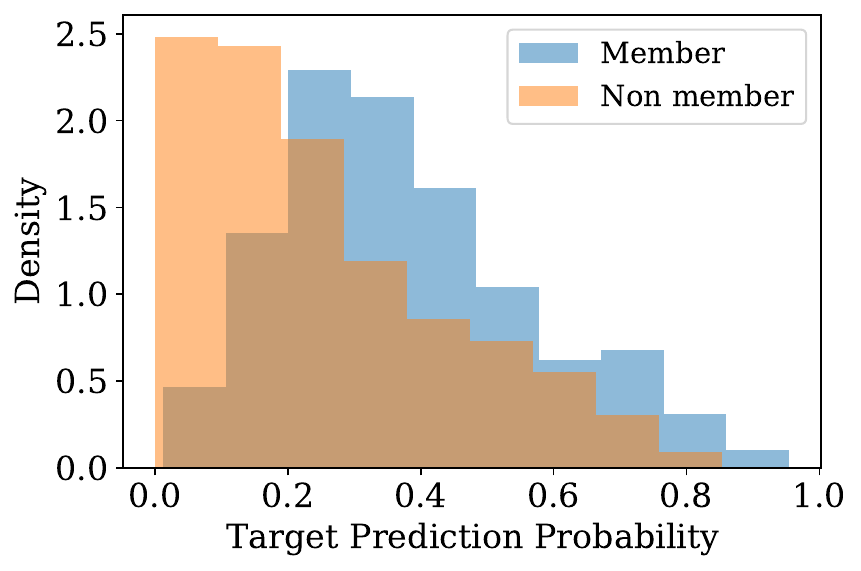}
            \caption[trec]%
            {{\small trec}}    
            \label{fig:mean and std of net44}
        \end{subfigure}
        \caption{\textbf{Output Probabilities at the Target Class for Members and Non-Members.} We depict the probability of the prompted GPT2-xl on the correct target class for member and non-member data points.}
        \label{fig:distribution_probs}
    \end{figure*}

\begin{figure*}
        \begin{subfigure}[b]{0.225\textwidth}
            \centering
            \includegraphics[width=\textwidth]{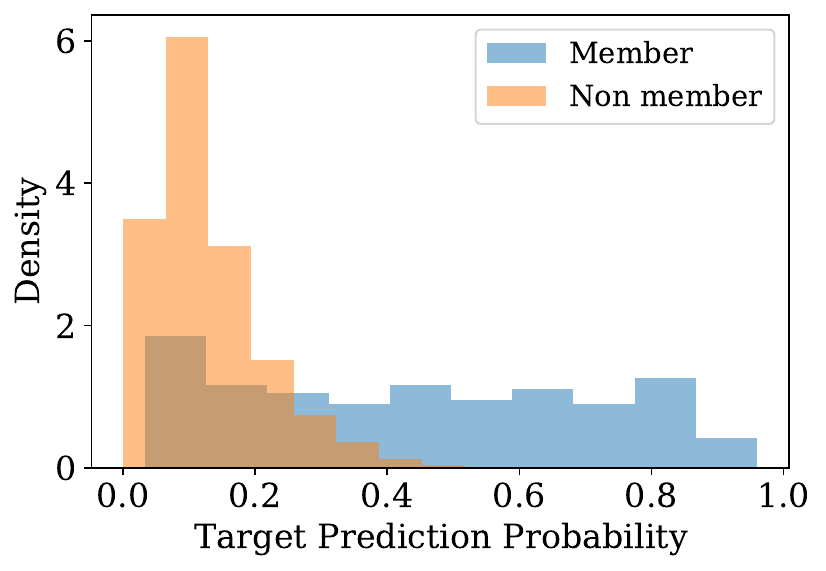}
            \caption[]%
            {{\small agnews}}    
            \label{fig:mean and std of net14}
        \end{subfigure}
        \hfill
        \begin{subfigure}[b]{0.225\textwidth}  
            \centering 
            \includegraphics[width=\textwidth]{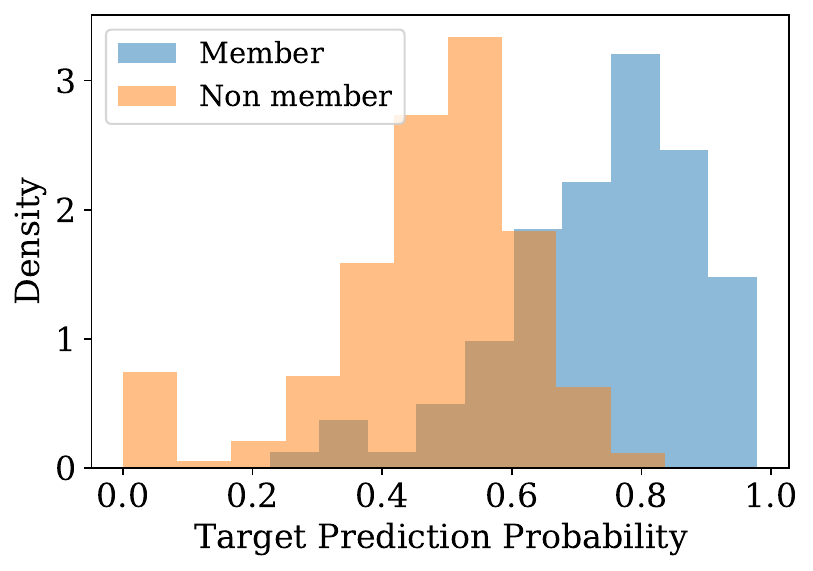}
            \caption[cb]%
            {{\small cb}}    
            \label{fig:mean and std of net24}
        \end{subfigure}
        \hfill
        \begin{subfigure}[b]{0.225\textwidth}   
            \centering 
            \includegraphics[width=\textwidth]{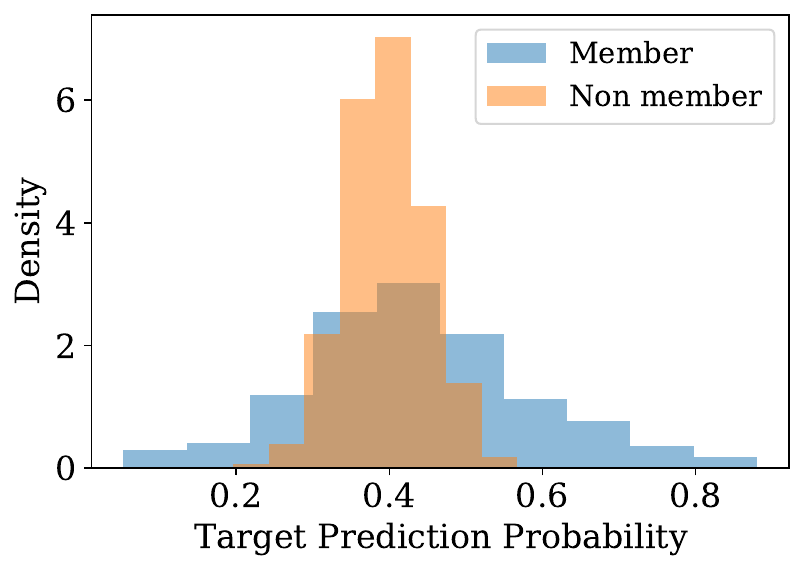}
            \caption[sst2]%
            {{\small sst2}}    
            \label{fig:mean and std of net34}
        \end{subfigure}
        \hfill
        \begin{subfigure}[b]{0.225\textwidth}   
            \centering 
            \includegraphics[width=\textwidth]{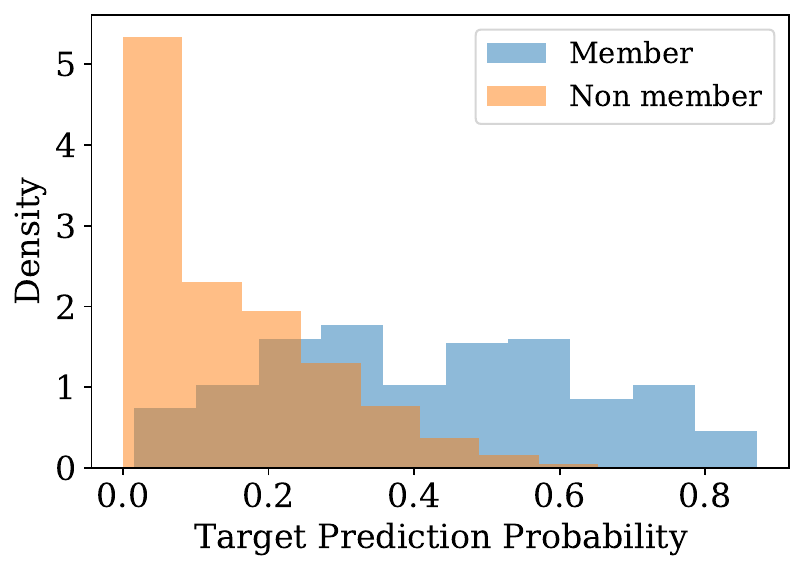}
            \caption[trec]%
            {{\small trec}}    
            \label{fig:mean and std of net44}
        \end{subfigure}
        \caption{\textbf{Output Probabilities at the Target Class for Members and Non-Members for GPT2-base.} We depict the probability of the ensemble of prompted GPT2-base on the correct target class for member and non-member data points.}
        \label{fig:distribution_base}
    \end{figure*}

\begin{figure*}
        \begin{subfigure}[b]{0.225\textwidth}
            \centering
            \includegraphics[width=\textwidth]{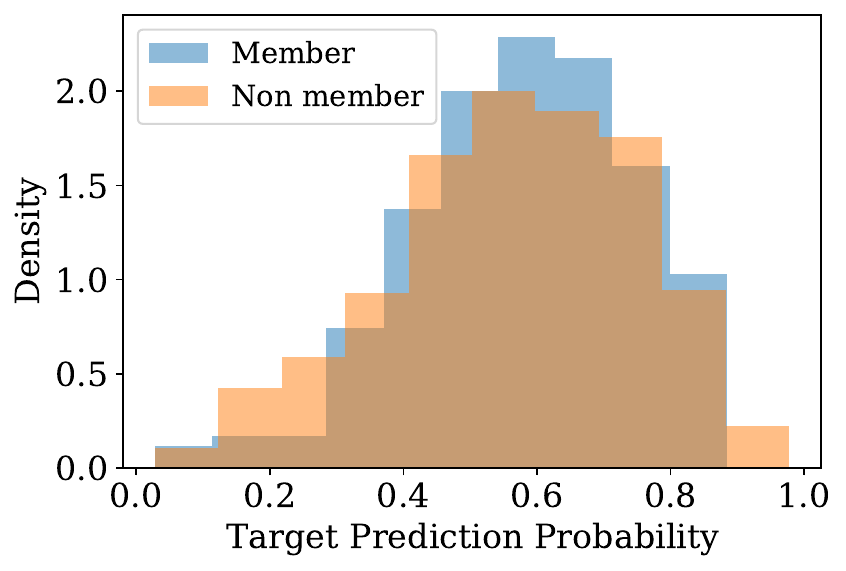}
            \caption[]%
            {{\small agnews}}    
            \label{fig:mean and std of net14}
        \end{subfigure}
        \hfill
        \begin{subfigure}[b]{0.225\textwidth}  
            \centering 
            \includegraphics[width=\textwidth]{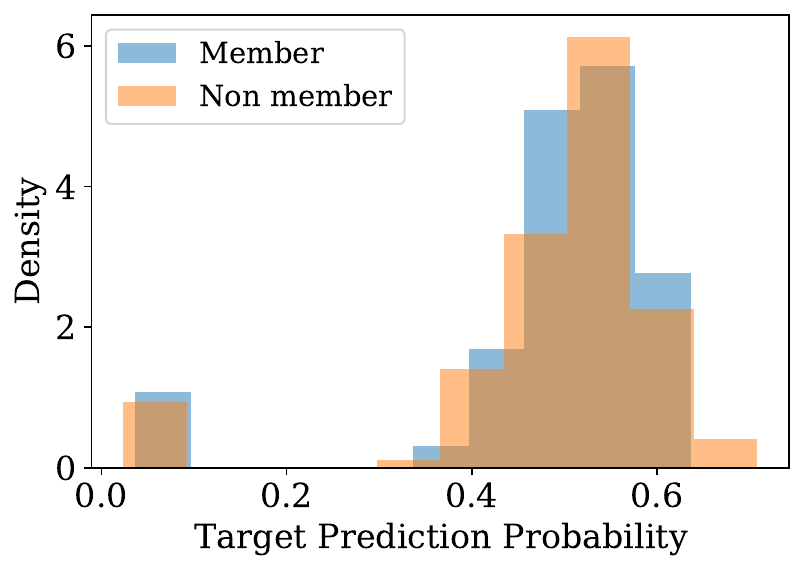}
            \caption[cb]%
            {{\small cb}}    
            \label{fig:mean and std of net24}
        \end{subfigure}
        \hfill
        \begin{subfigure}[b]{0.225\textwidth}   
            \centering 
            \includegraphics[width=\textwidth]{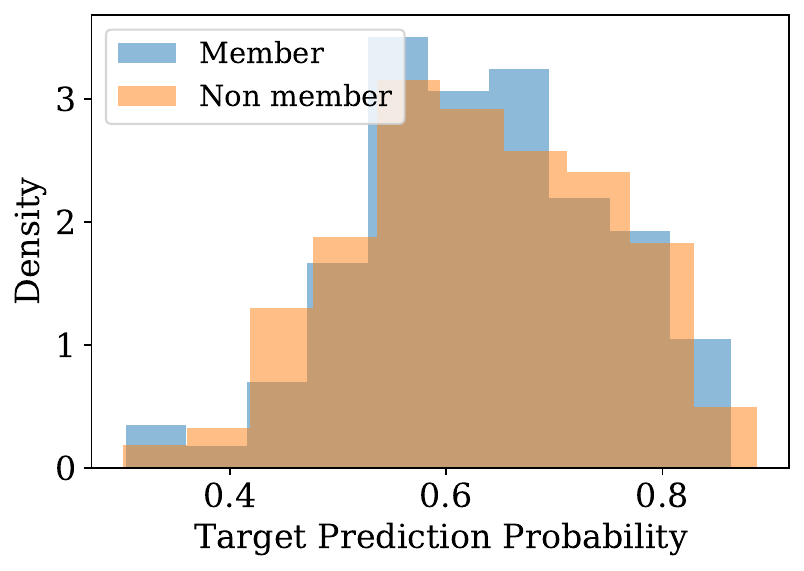}
            \caption[sst2]%
            {{\small sst2}}    
            \label{fig:mean and std of net34}
        \end{subfigure}
        \hfill
        \begin{subfigure}[b]{0.225\textwidth}   
            \centering 
            \includegraphics[width=\textwidth]{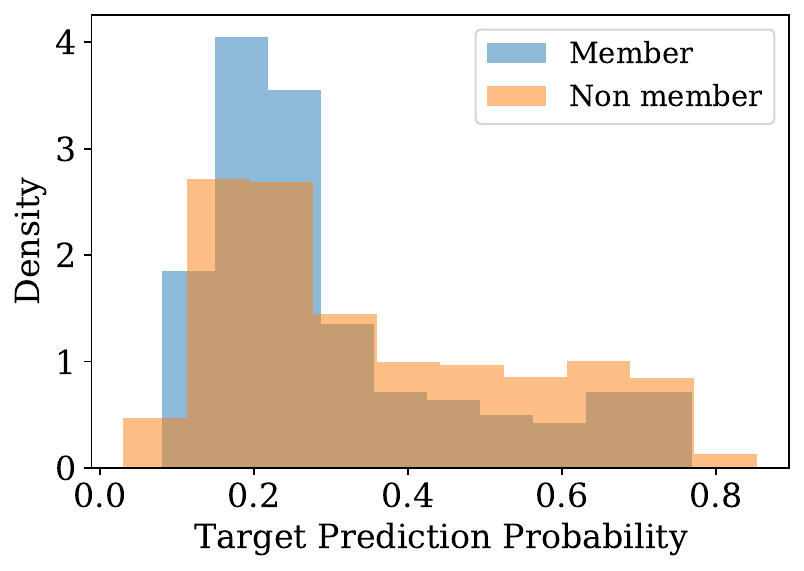}
            \caption[trec]%
            {{\small trec}}    
            \label{fig:mean and std of net44}
        \end{subfigure}
        \caption{\textbf{Output Probabilities at the Target Class for Members and Non-Members under \nameA.} We depict the probability of the ensemble of prompted GPT2-xl on the correct target class for member and non-member data points. We perform ensembling by aggregating the raw output probabilities over \numEns prompted models and computing the average output vector. We find that the discrepancy between member and non-member becomes much smaller after ensembling. }
        \label{fig:distribution_ensembling}
    \end{figure*}

\begin{figure*}
        \begin{subfigure}[b]{0.225\textwidth}
            \centering
            \includegraphics[width=\textwidth]{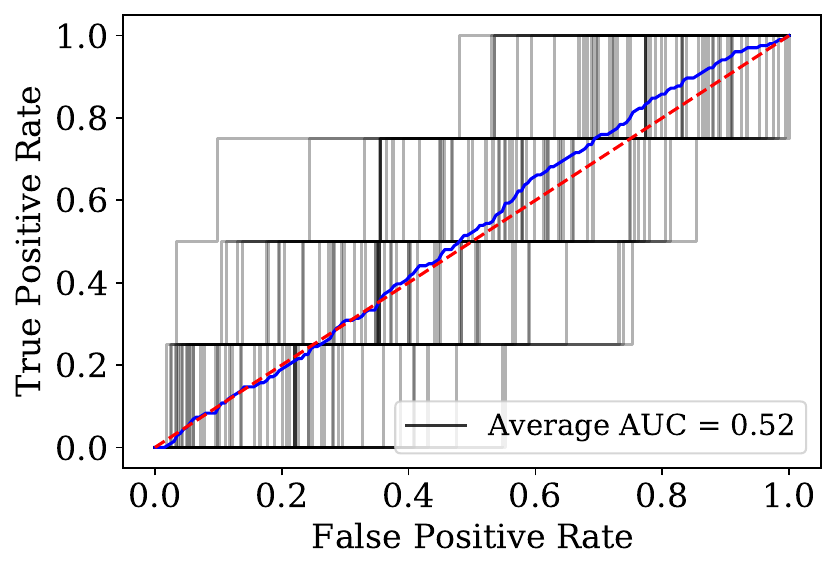}
            \caption[]%
            {{\small agnews}}    
            \label{fig:mean and std of net14}
        \end{subfigure}
        \hfill
        \begin{subfigure}[b]{0.225\textwidth}  
            \centering 
            \includegraphics[width=\textwidth]{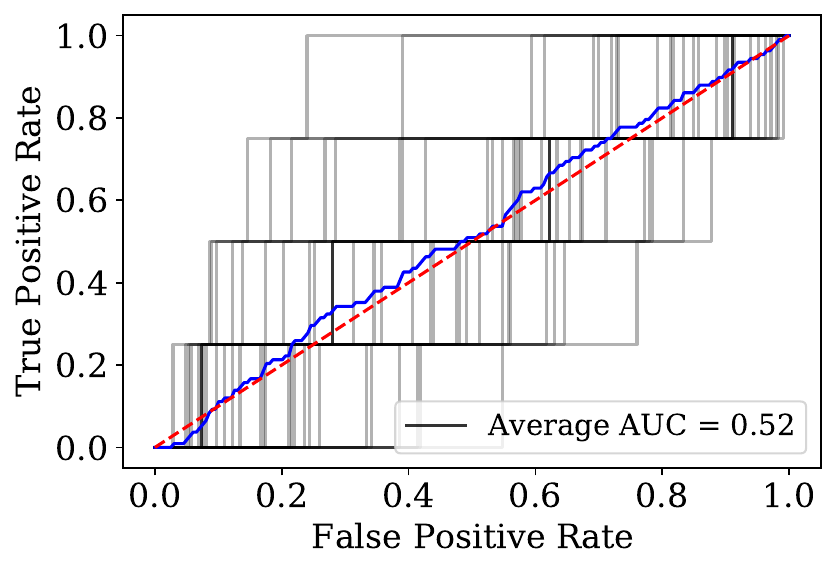}
            \caption[cb]%
            {{\small cb}}    
            \label{fig:mean and std of net24}
        \end{subfigure}
        \hfill
        \begin{subfigure}[b]{0.225\textwidth}   
            \centering 
            \includegraphics[width=\textwidth]{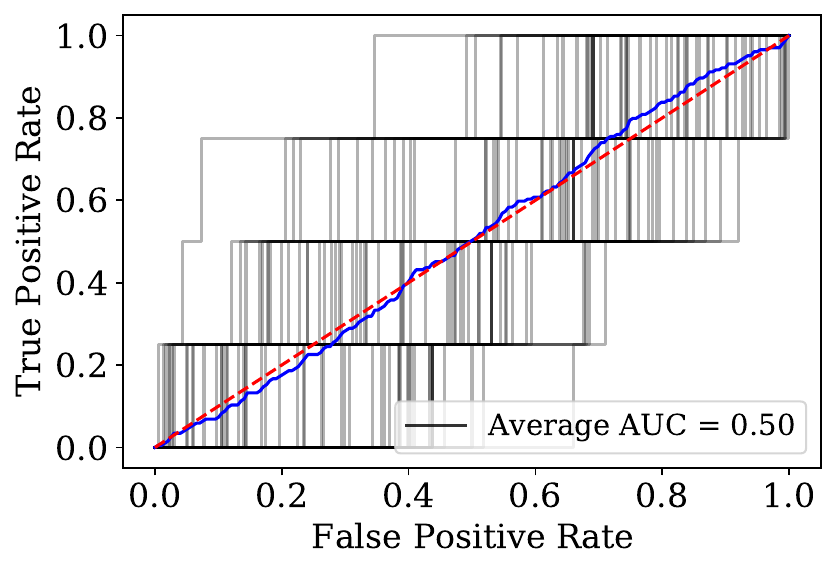}
            \caption[sst2]%
            {{\small sst2}}    
            \label{fig:mean and std of net34}
        \end{subfigure}
        \hfill
        \begin{subfigure}[b]{0.225\textwidth}   
            \centering 
            \includegraphics[width=\textwidth]{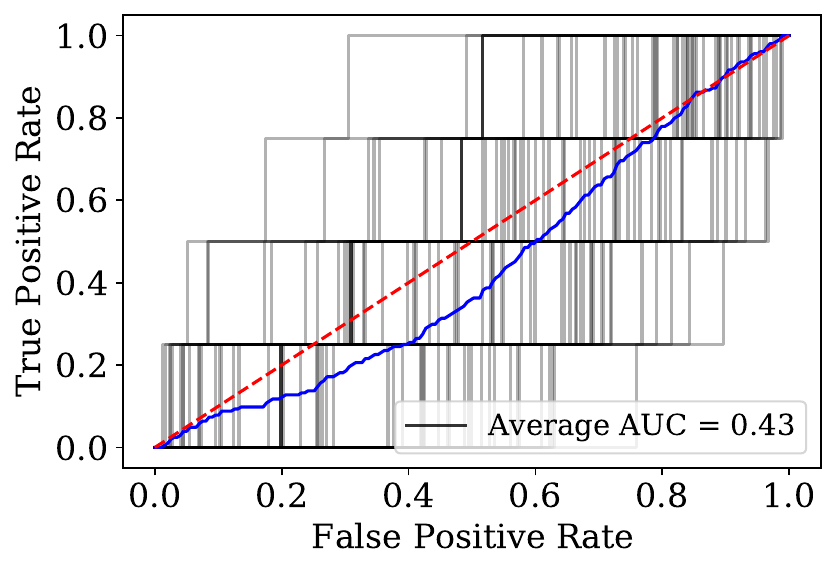}
            \caption[trec]%
            {{\small trec}}    
            \label{fig:mean and std of net44}
        \end{subfigure}
        \caption{\textbf{MIA risk over all Datasets after (\nameA).} We depict the AUC-ROC of MIA after \nameA. Across all datasets, the effectiveness of MIA (blue line) is close to random guessing (red line).}
        \label{fig:mi_attack_ensembling}
    \end{figure*}

\begin{figure*}
        \begin{subfigure}[b]{0.225\textwidth}
            \centering
            \includegraphics[width=\textwidth]{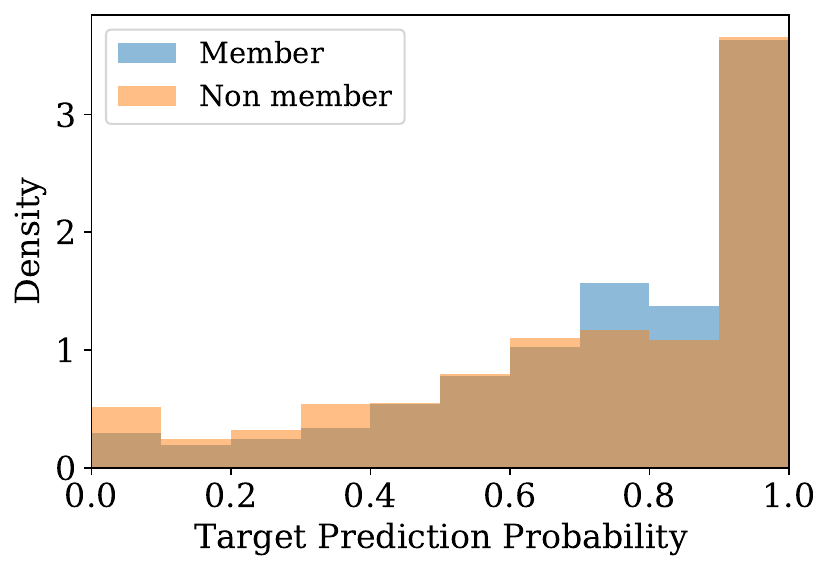}
            \caption[]%
            {{\small agnews}}    
            \label{fig:mean and std of net14}
        \end{subfigure}
        \hfill
        \begin{subfigure}[b]{0.225\textwidth}  
            \centering 
            \includegraphics[width=\textwidth]{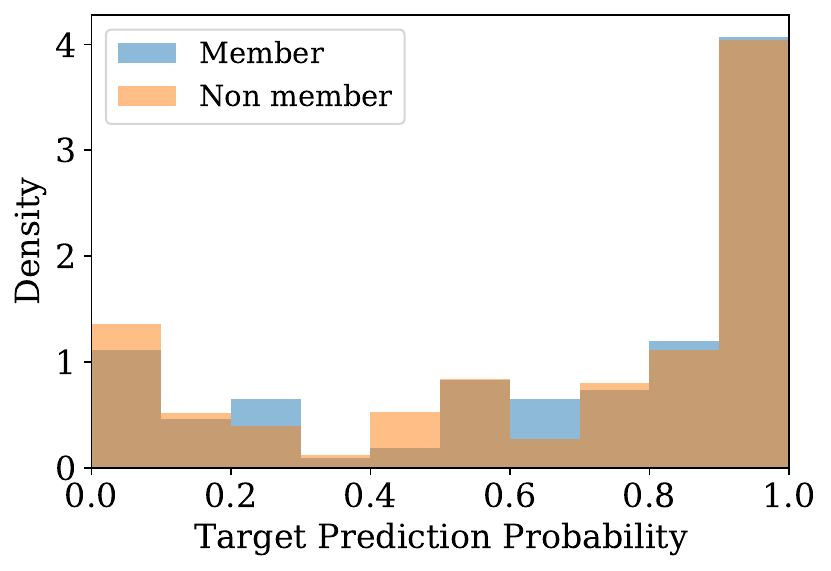}
            \caption[cb]%
            {{\small cb}}    
            \label{fig:mean and std of net24}
        \end{subfigure}
        \hfill
        \begin{subfigure}[b]{0.225\textwidth}   
            \centering 
            \includegraphics[width=\textwidth]{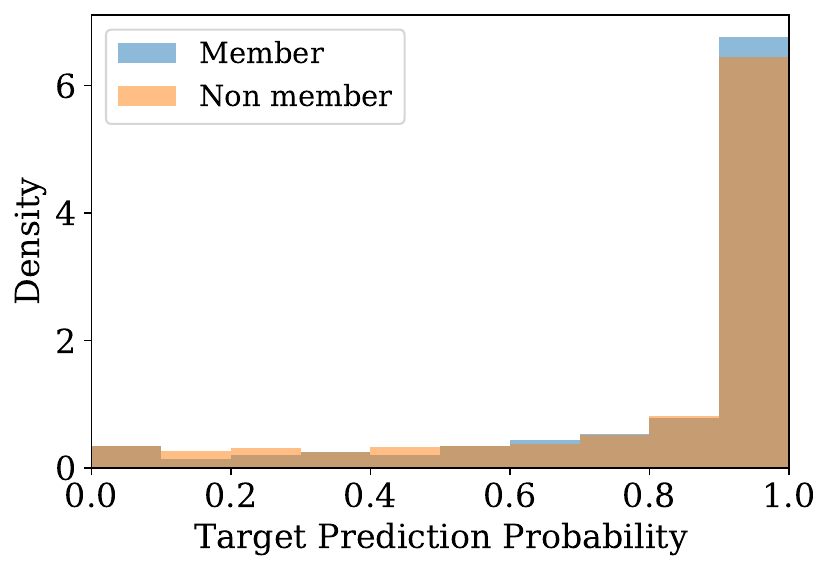}
            \caption[sst2]%
            {{\small sst2}}    
            \label{fig:mean and std of net34}
        \end{subfigure}
        \hfill
        \begin{subfigure}[b]{0.225\textwidth}   
            \centering 
            \includegraphics[width=\textwidth]{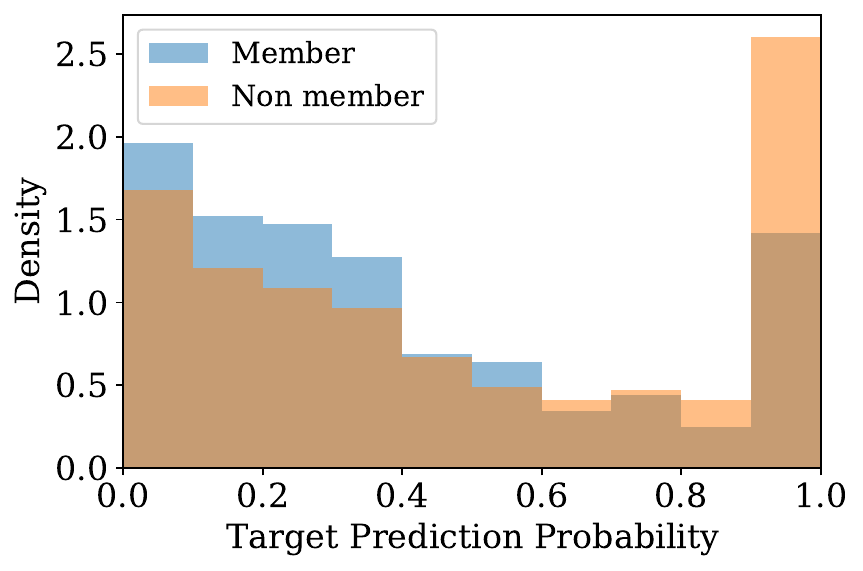}
            \caption[trec]%
            {{\small trec}}    
            \label{fig:mean and std of net44}
        \end{subfigure}
        \caption{\textbf{Voting Probabilities for the Correct Target Class with \nameB.} We ensemble the individual prompted models by obtaining the class with the highest prediction probability from each model. We show for member and non-member data points what percentage of the \numEns prompted models returns the correct target class. This corresponds to the confidence of the ensemble.}
        \label{fig:distribution_vote}
    \end{figure*}

\begin{figure*}
        \begin{subfigure}[b]{0.225\textwidth}
            \centering
            \includegraphics[width=\textwidth]{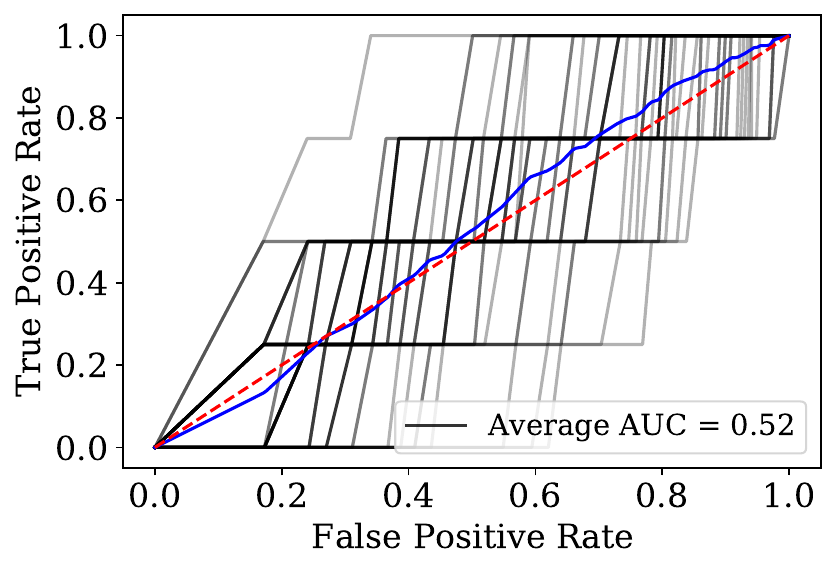}
            \caption[]%
            {{\small agnews}}    
            \label{fig:mean and std of net14}
        \end{subfigure}
        \hfill
        \begin{subfigure}[b]{0.225\textwidth}  
            \centering 
            \includegraphics[width=\textwidth]{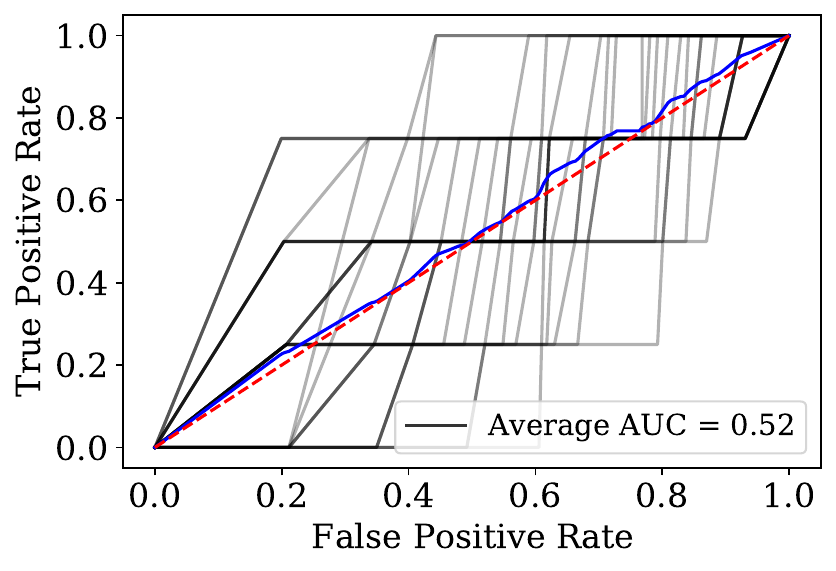}
            \caption[cb]%
            {{\small cb}}    
            \label{fig:mean and std of net24}
        \end{subfigure}
        \hfill
        \begin{subfigure}[b]{0.225\textwidth}   
            \centering 
            \includegraphics[width=\textwidth]{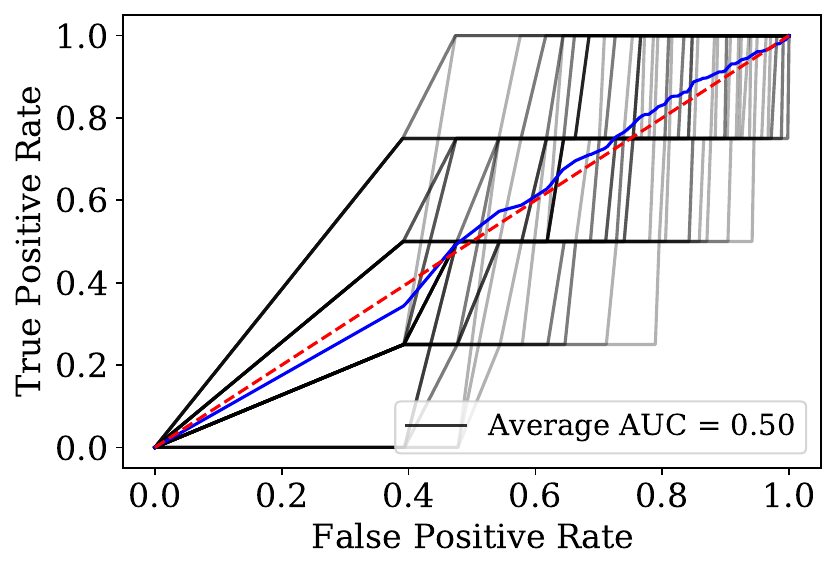}
            \caption[sst2]%
            {{\small sst2}}    
            \label{fig:mean and std of net34}
        \end{subfigure}
        \hfill
        \begin{subfigure}[b]{0.225\textwidth}   
            \centering 
            \includegraphics[width=\textwidth]{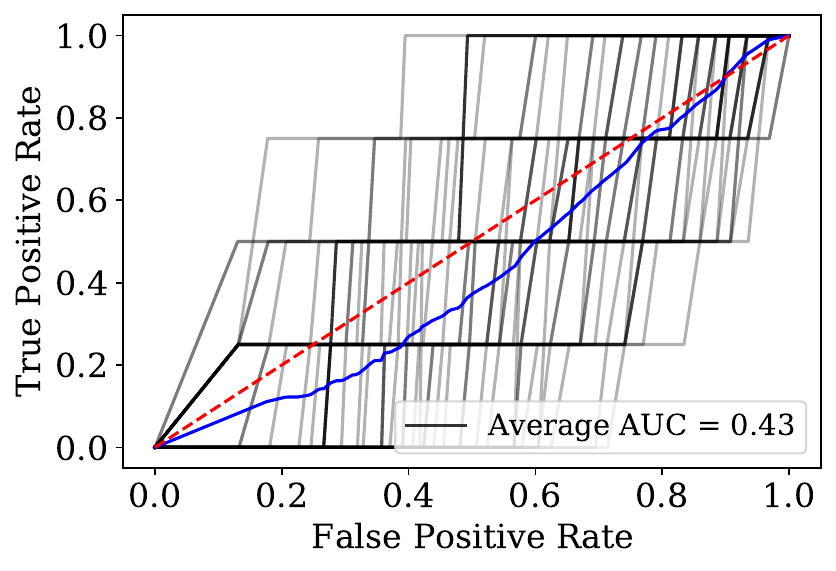}
            \caption[trec]%
            {{\small trec}}    
            \label{fig:mean and std of net44}
        \end{subfigure}
        \caption[ The average and standard deviation of critical parameters ]
        {\textbf{MIA Risk over all Datasets (\nameB).} We depict the AUC-ROC of MIA after \nameB. Across all datasets, the effectiveness of MIA (blue line) is close to random guessing (red line).}
        \label{fig:mi_attack_vote}
    \end{figure*}

\begin{figure*}
        \begin{subfigure}[b]{0.225\textwidth}
            \centering
            \includegraphics[width=\textwidth]{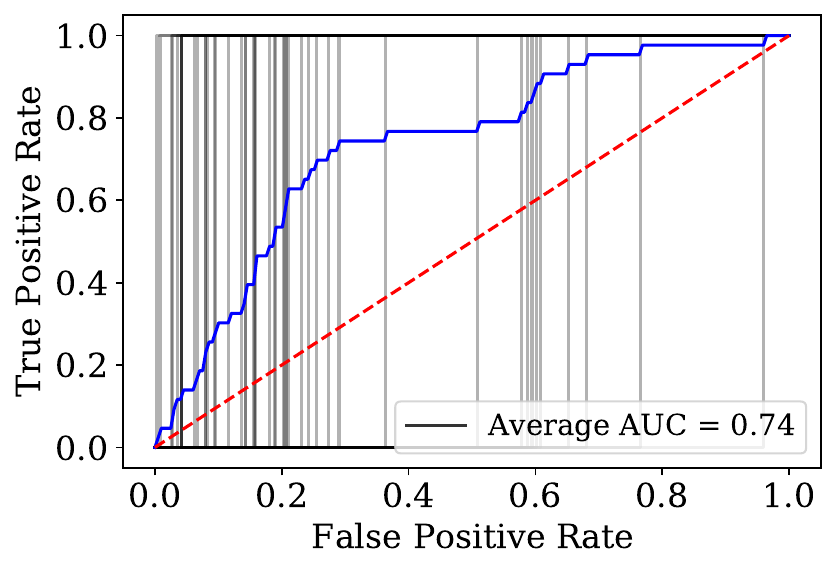}
            \caption[]%
            {{\small agnews}}    
            \label{fig:mean and std of net14}
        \end{subfigure}
        \hfill
        \begin{subfigure}[b]{0.225\textwidth}  
            \centering 
            \includegraphics[width=\textwidth]{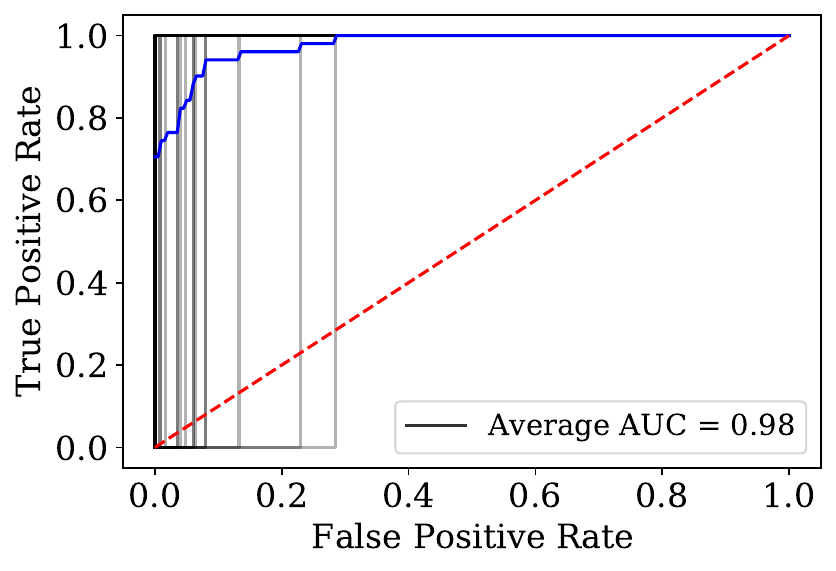}
            \caption[cb]%
            {{\small cb}}    
            \label{fig:mean and std of net24}
        \end{subfigure}
        \hfill
        \begin{subfigure}[b]{0.225\textwidth}   
            \centering 
            \includegraphics[width=\textwidth]{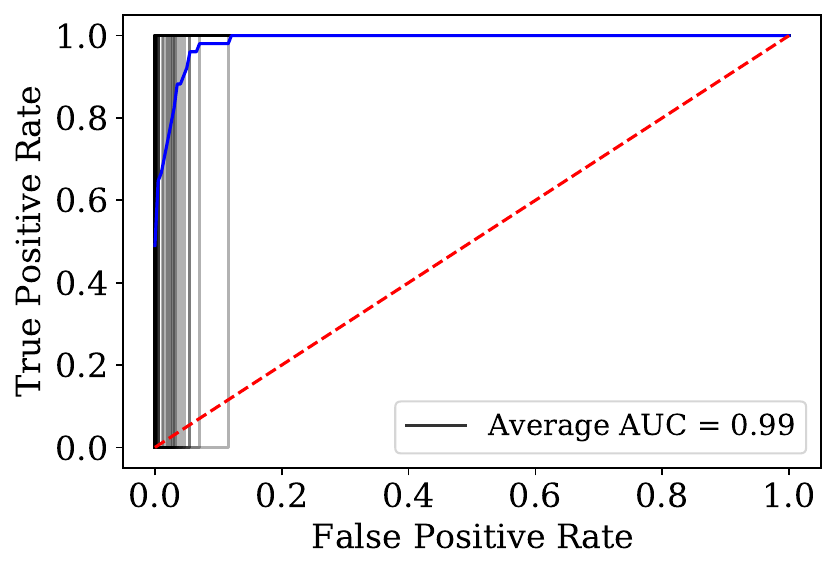}
            \caption[sst2]%
            {{\small sst2}}    
            \label{fig:mean and std of net34}
        \end{subfigure}
        \hfill
        \begin{subfigure}[b]{0.225\textwidth}   
            \centering 
            \includegraphics[width=\textwidth]{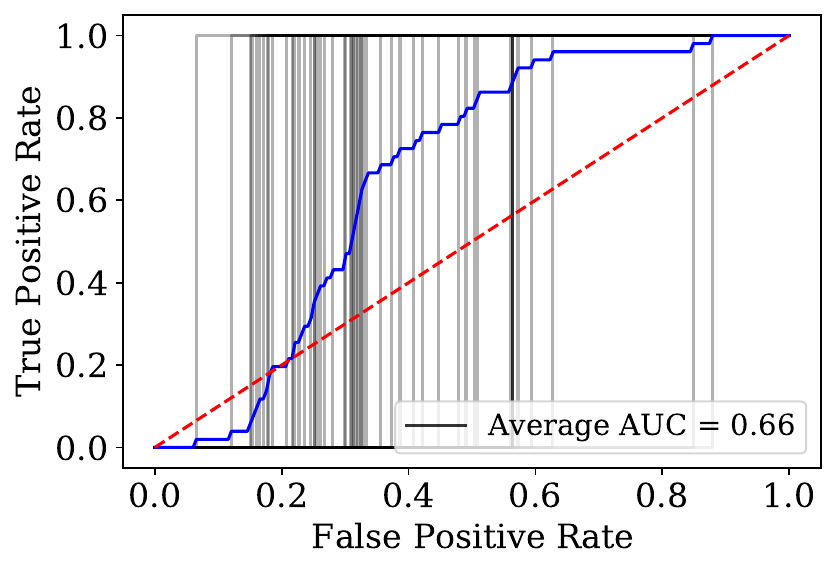}
            \caption[trec]%
            {{\small trec}}    
            \label{fig:mean and std of net44}
        \end{subfigure}
        \caption{\textbf{MIA Risk over all Datasets for One-Shot Learning.} This figure  corresponds to \Cref{fig:mi_attack} with the difference that we only use one example (instead of four) in the prompt. We depict the AUC-ROC curves over all datasets. The red dashed line
represents the MIA success of random guessing. Each gray line corresponds to a prompted model with its four member data points. Due to the small number of member data points (1), our resulting TPRs can only be 0\% or 100\% which leads to the step-shape of the gray curves.
        The reported average AUC-score is calculated as an average over the individual prompted models (gray lines)' AUC score.
        Additionally, for visualization purposes, we average the gray lines over all prompted models and depict the average as the blue line.
        We use \numChosen prompted models in this experiment..}
        \label{fig:mi_attack_ensembling}
    \end{figure*}

\begin{figure*}
        \begin{subfigure}[b]{0.225\textwidth}
            \centering
            \includegraphics[width=\textwidth]{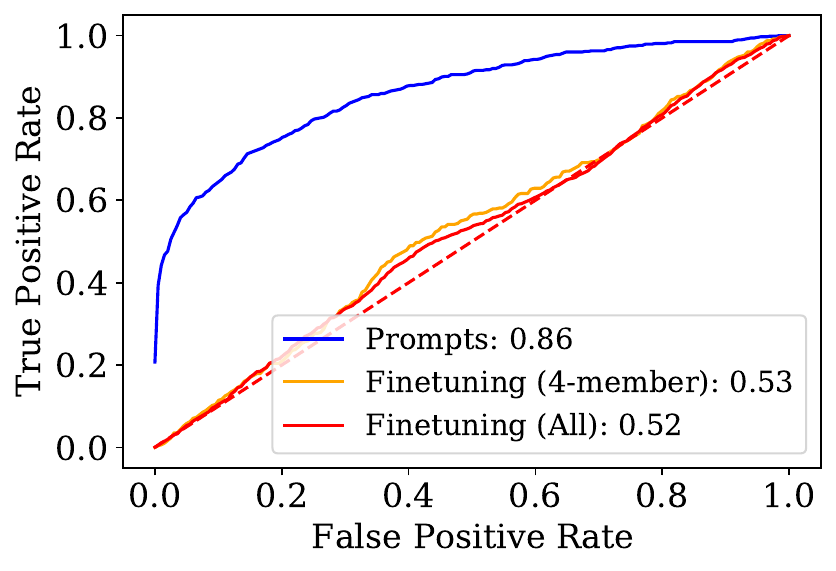}
            \caption[]%
            {{\small agnews}}    
            \label{fig:mean and std of net14}
        \end{subfigure}
        \hfill
        \begin{subfigure}[b]{0.225\textwidth}  
            \centering 
            \includegraphics[width=\textwidth]{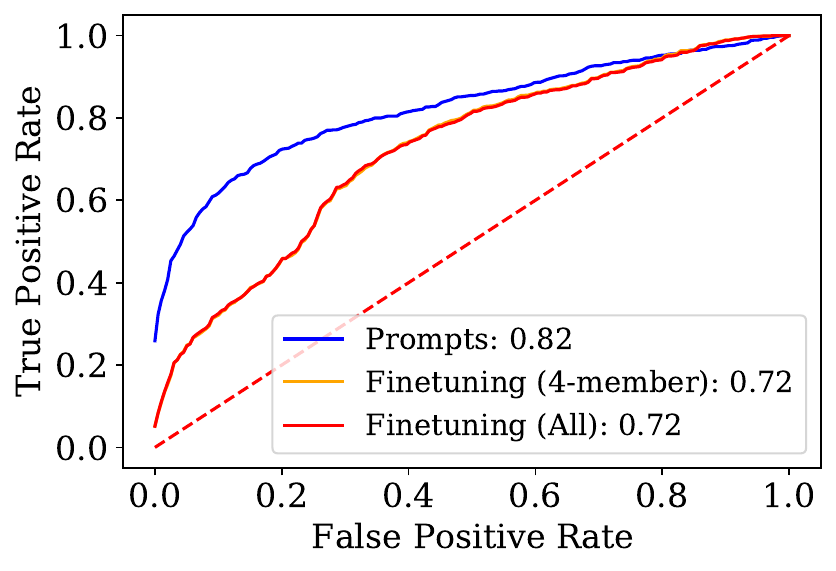}
            \caption[cb]%
            {{\small cb}}    
            \label{fig:mean and std of net24}
        \end{subfigure}
        \hfill
        \begin{subfigure}[b]{0.225\textwidth}   
            \centering 
            \includegraphics[width=\textwidth]{figure/sst2/finetune_auc_roc_64.pdf}
            \caption[sst2]%
            {{\small sst2}}    
            \label{fig:mean and std of net34}
        \end{subfigure}
        \hfill
        \begin{subfigure}[b]{0.225\textwidth}   
            \centering 
            \includegraphics[width=\textwidth]{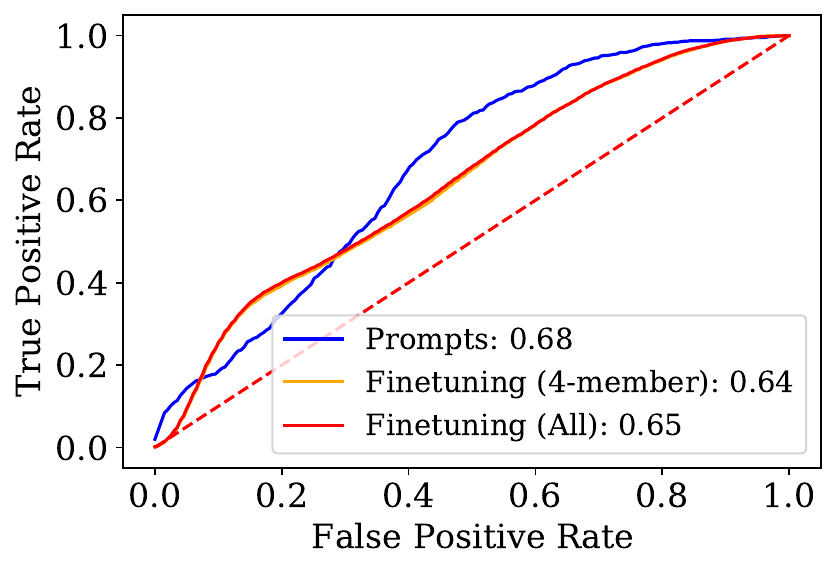}
            \caption[trec]%
            {{\small trec}}    
            \label{fig:mean and std of net44}
        \end{subfigure}
        \caption{\textbf{MIA on Fine-Tuning vs Prompting across all Datasets.} We plot our MIA risk on prompted and fine-tuned models given similar downstream performance. For fine-tuning, we evaluate MIA risk in two different ways to avoid the influence of different training set size. The red dashed line represents the MIA success of random guessing. The results show that prompts are much more vulnerable to MIA than fine-tuning.}
        \label{fig:append_fine_tune_auc}
    \end{figure*}

\begin{figure*}
        \begin{subfigure}[b]{0.225\textwidth}
            \centering
            \includegraphics[width=\textwidth]{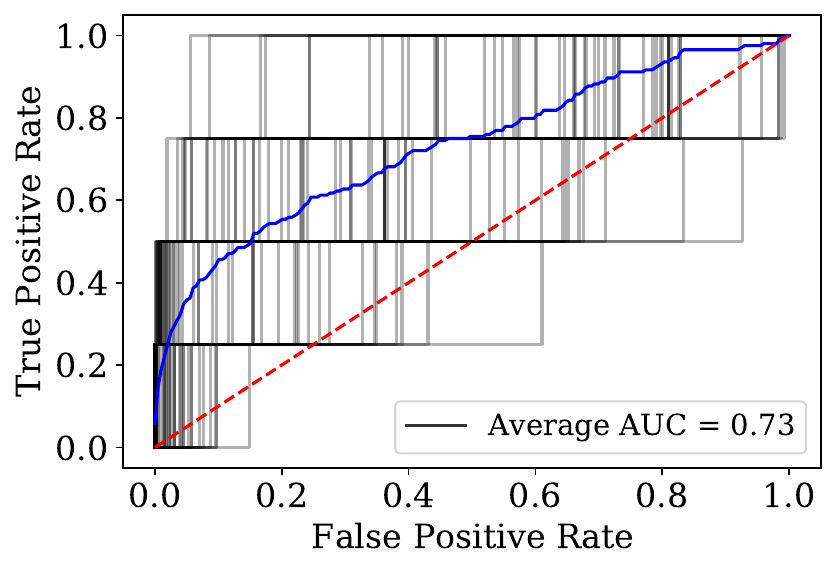}
            \caption[]%
            {{\small agnews: normalized}}    
            \label{fig:mean and std of net14}
        \end{subfigure}
        \hfill
        \begin{subfigure}[b]{0.225\textwidth}  
            \centering 
            \includegraphics[width=\textwidth]{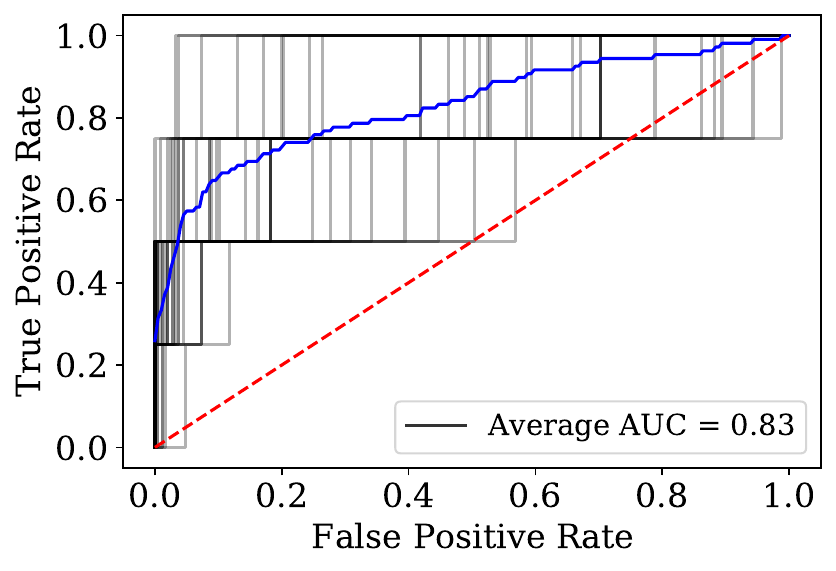}
            \caption[cb]%
            {{\small cb: normalized}}    
            \label{fig:mean and std of net24}
        \end{subfigure}
        \begin{subfigure}[b]{0.225\textwidth}   
            \centering 
            \includegraphics[width=\textwidth]{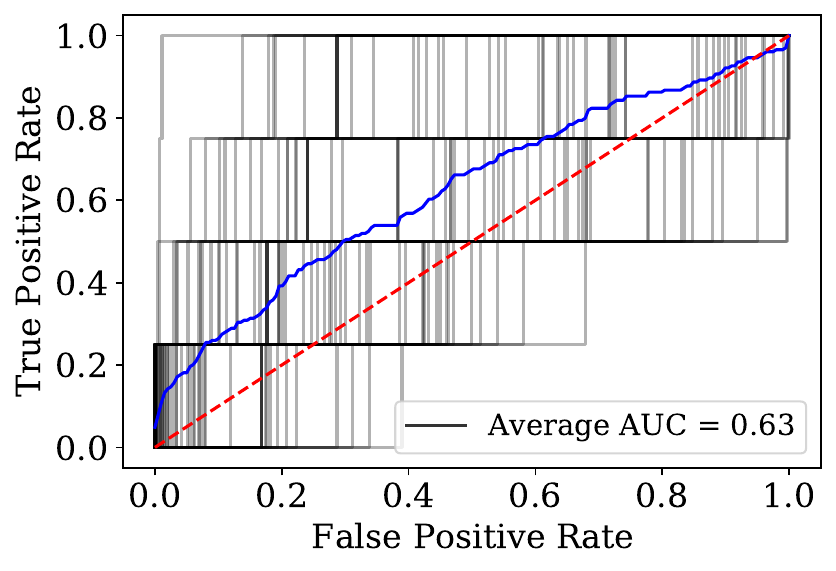}
            \caption[sst2]%
            {{\small sst2: normalized}}    
            \label{fig:mean and std of net34}
        \end{subfigure}
        \hfill
        \begin{subfigure}[b]{0.225\textwidth}   
            \centering 
            \includegraphics[width=\textwidth]{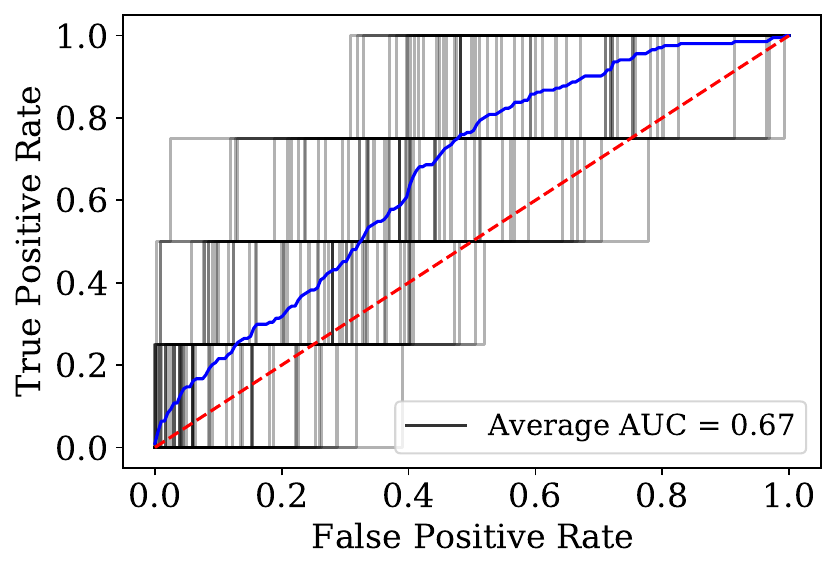}
            \caption[rte]%
            {{\small trec: normalized}}    
            \label{fig:mean and std of net44}
        \end{subfigure}
        \caption{\textbf{Impact of Normalization.} We report the AUC for our MIA on prompted GPT2-xl for normalized outputs, \ie outputs where the probabilities over all target classes of the respective downstream task add up to one.}
        \label{fig:mi_attack_normalized}
    \end{figure*}

\begin{figure*}
        \begin{subfigure}[b]{0.225\textwidth}
            \centering
            \includegraphics[width=\textwidth]{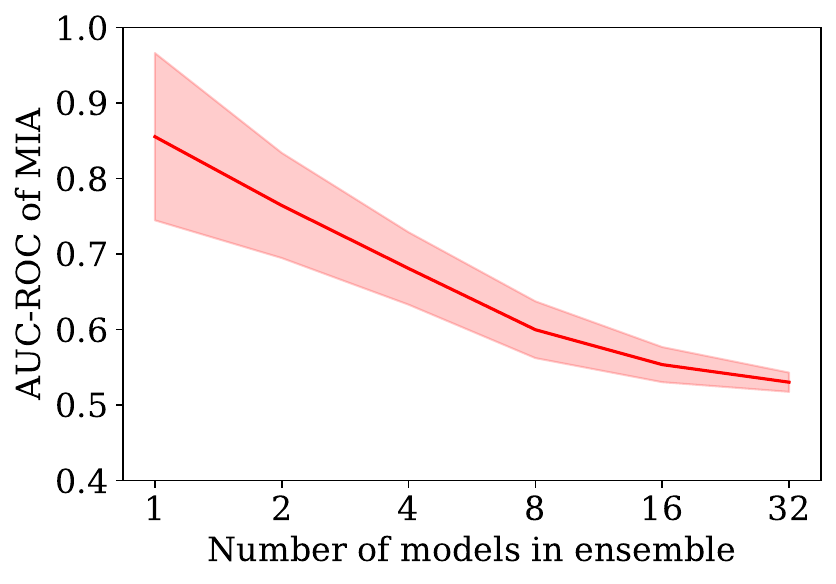}
            \caption[]%
            {{\small agnews }}    
            \label{fig:mean and std of net14}
        \end{subfigure}
        \hfill
        \begin{subfigure}[b]{0.225\textwidth}  
            \centering 
            \includegraphics[width=\textwidth]{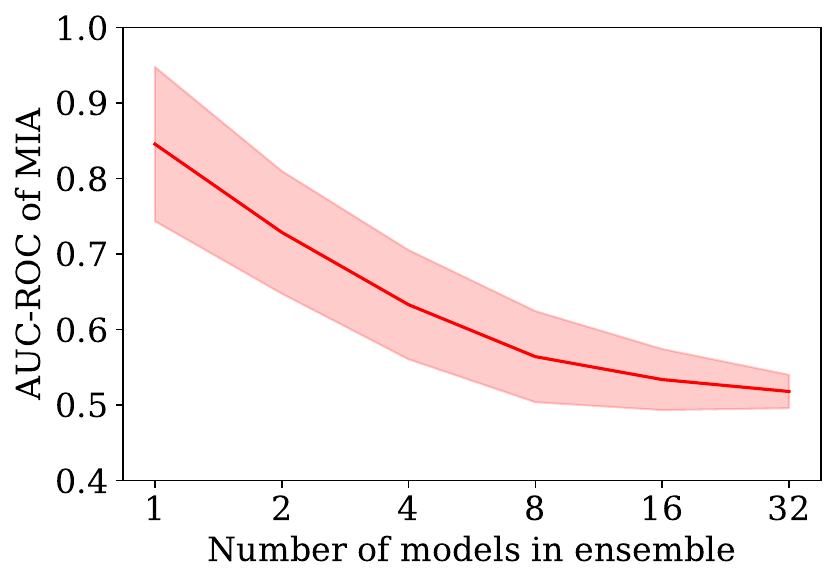}
            \caption[cb]%
            {{\small cb}}    
            \label{fig:mean and std of net24}
        \end{subfigure}
        \hfill
        \begin{subfigure}[b]{0.225\textwidth}   
            \centering 
            \includegraphics[width=\textwidth]{figure/number_of_teachers_sst2.pdf}
            \caption[sst2]%
            {{\small sst2}}    
            \label{fig:mean and std of net34}
        \end{subfigure}
        \hfill
        \begin{subfigure}[b]{0.225\textwidth}   
            \centering 
            \includegraphics[width=\textwidth]{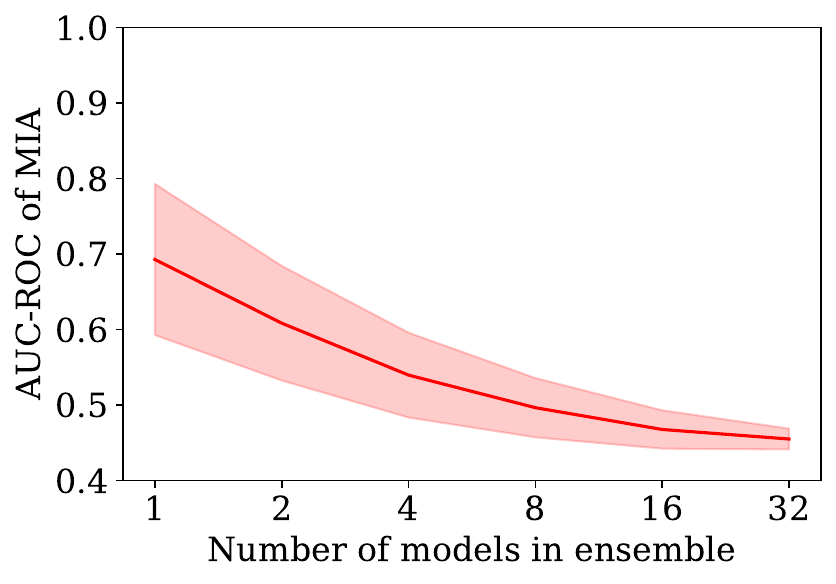}
            \caption[trec]%
            {{\small trec}}    
            \label{fig:mean and std of net44}
        \end{subfigure}
        \caption{\textbf{Number of teachers in average ensemble vs MIA risks.}  We plot the membership risk in form of the AUC score of MIA while we vary the number of teachers in ensembling. We observe that with more data used for ensembling, the lower risk of MIA (in terms of AUC and its variance).}
        \label{fig:append_number_of_teachers}
    \end{figure*}

\end{document}